\documentclass{article}

\usepackage[final]{neurips_2022}
\usepackage[utf8]{inputenc}
\usepackage[T1]{fontenc}
\usepackage{hyperref}
\usepackage{url}
\usepackage{booktabs}
\usepackage{amsfonts}
\usepackage{nicefrac}
\usepackage{microtype}
\usepackage{xcolor}

\usepackage{amsfonts}
\usepackage{amsmath}
\usepackage{amssymb}
\usepackage{comment}
\usepackage{float}
\usepackage{listings}
\usepackage{mathtools}
\usepackage{subcaption}

\newcommand{\R}{\mathbb{R}}
\newcommand{\N}{\mathcal{N}}
\newcommand{\U}{\mathcal{U}}
\newcommand{\X}{\mathcal{X}}
\newcommand{\Y}{\mathcal{Y}}

\renewcommand{\b}[1]{\mathbf{#1}}

\DeclareMathOperator*{\argmax}{\arg\!\max}
\DeclareMathOperator{\sgn}{sgn}
\DeclareMathOperator{\E}{\mathbb{E}}
\DeclareMathOperator{\Cov}{Cov}

\title{Generative multitask learning mitigates target-causing confounding}

\author{
Taro Makino$^{1}$\quad
Krzysztof J. Geras$^{2,1}$\quad
Kyunghyun Cho$^{1,3,4}$
\\\\
$^1$NYU Center for Data Science\\
$^2$NYU Grossman School of Medicine\\
$^3$Genentech \enspace $^4$CIFAR LMB\\
\texttt{taro@nyu.edu}
}

\begin{document}

\maketitle

\begin{abstract}
We propose generative multitask learning (GMTL), a simple and scalable approach to causal representation learning for multitask learning. Our approach makes a minor change to the conventional multitask inference objective, and improves robustness to target shift. Since GMTL only modifies the inference objective, it can be used with existing multitask learning methods without requiring additional training. The improvement in robustness comes from mitigating unobserved confounders that cause the targets, but not the input. We refer to them as \emph{target-causing confounders}. These confounders induce spurious dependencies between the input and targets. This poses a problem for conventional multitask learning, due to its assumption that the targets are conditionally independent given the input. GMTL mitigates target-causing confounding at inference time, by removing the influence of the joint target distribution, and predicting all targets jointly. This removes the spurious dependencies between the input and targets, where the degree of removal is adjustable via a single hyperparameter. This flexibility is useful for managing the trade-off between in- and out-of-distribution generalization. Our results on the Attributes of People and Taskonomy datasets reflect an improved robustness to target shift across four multitask learning methods.
\end{abstract}

\section{Introduction}

Deep neural networks (DNNs) excel at extracting patterns from unstructured data, but these patterns can fail to generalize outside of the training distribution. These failures are often attributed to DNNs learning statistical associations rather than causal relations~\citep{Ribeiro2016WhyShouldITrustYou,Jo2017SurfaceStatisticalRegularities,Geirhos2020ShortcutLearning}. Therefore, there is a concerted effort to make patterns learned by DNNs satisfy certain properties of causal relations, such as invariance and modularity~\citep{Scholkopf2019CausalityML,Scholkopf2021CRL}. This research direction is called causal representation learning. Existing approaches to causal representation learning often require additional information in the form of labeled environments~\citep{Arjovsky2019IRM,Lu2021NonlinearIRM} or labeled confounders~\citep{Puli2021NuisanceSpuriousCorrelations}. These requirements are restrictive, and it is therefore beneficial to identify more natural settings where existing information can be used for causal representation learning.

\begin{figure}[t]
    \centering
    \hspace*{\fill}
    \begin{subfigure}[t]{0.2\textwidth}
        \centering
        \includegraphics[width=\textwidth]{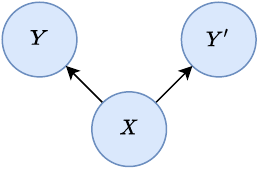}
        \caption{DMTL}
        \label{fig:dmtl}
    \end{subfigure}
    \hfill
    \begin{subfigure}[t]{0.2\textwidth}
        \centering
        \includegraphics[width=\textwidth]{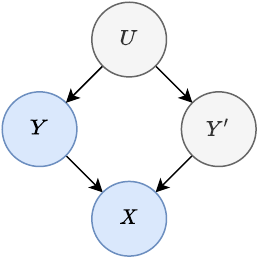}
        \caption{DMTL (alternative assumptions)}
        \label{fig:backdoor path}
    \end{subfigure}
    \hfill
    \begin{subfigure}[t]{0.2\textwidth}
        \centering
        \includegraphics[width=\textwidth]{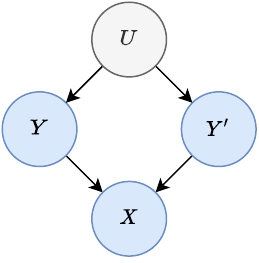}
        \caption{GMTL}
        \label{fig:gmtl}
    \end{subfigure}
    \hspace*{\fill}
    \caption{(a) The conventional approach to MTL, which we call discriminative multitask learning (DMTL), assumes the targets $\b{y}$ and $\b{y}'$ are conditionally independent given the input $\b{x}$. Due to this assumption, the inference objective for predicting the main target $\b{y}$ is $\argmax_\b{y} \log p(\b{y} \mid \b{x})$. (b) DMTL is flawed under an alternative set of assumptions, where the targets cause the input, and there exists an unobserved confounder $\b{u}$ that causes the targets, but not the input. We call $\b{u}$ a target-causing confounder. Since $\b{y}'$ is unobserved when predicting $\b{y}$ with DMTL, this opens up the backdoor path $\b{y} \leftarrow \b{u} \rightarrow \b{y}' \rightarrow \b{x}$ and makes the input and targets spuriously dependent. That is, $p(\b{y} \mid \b{x})$ shifts when $p(\b{u})$ shifts. (c) We propose generative multitask learning (GMTL), where the inference objective is $\argmax_{\b{y}, \b{y}'} \log p(\b{x} \mid \b{y}, \b{y}')$. Unlike DMTL, this objective conditions on all targets, which $d$-separates $\b{x}$ and $\b{u}$, and removes the spurious dependencies between the input and targets.}
\end{figure}

We consider the setting of multitask learning (MTL)~\citep{Caruana1997MTL,Ruder2017MTLOverview,Zhang2017MTLSurvey}, where there is an input $\b{x} \in \X$ and multiple targets. Without loss of generality, we consider two targets $\b{y} \in \Y$ and $\b{y}' \in \Y'$. As is common in the MTL literature, we distinguish $\b{y}$ as being the main target, and $\b{y}'$ as the auxiliary target. Both targets are used during training, but only the main target is used during evaluation. In the conventional approach to MTL, which we call \emph{discriminative multitask learning} (DMTL), the inference objective is
\begin{align}
    \label{eq:dmtl factorized}
    \argmax_{\b{y}, \b{y}'} \log p(\b{y}, \b{y}' \mid \b{x}) = \argmax_{\b{y}, \b{y}'} \log p(\b{y} \mid \b{x}) + \log p(\b{y}' \mid \b{x}).
\end{align}
This factorization corresponds to assuming the targets are conditionally independent given the input. This is a convenient assumption to make, since it prevents $\Y{\times}\Y'$ from growing exponentially with the number of tasks. Since only the main target $\b{y}$ is used during evaluation, the inference objective is
\begin{align}
    \label{eq:dmtl final}
    \argmax_\b{y} \log p(\b{y} \mid \b{x}).
\end{align}
DMTL does not make causal assumptions explicitly, but its conditional independence assumption is consistent with the causal graph in Fig.~\ref{fig:dmtl}. We argue that DMTL is flawed under an alternative set of assumptions.

First, we assume the targets cause the input, which implies we are predicting the cause from the effect. This is called anticausal learning, and is considered to describe many common problems such as image classification~\citep{Scholkopf2012CausalAnticausal}. For example, in object recognition, the object category causes the pixel representation of that object. Second, we assume there exists a variable $\b{u}$ that causes the targets, but not the input. We call $\b{u}$ a \emph{target-causing confounder}.

The problem with DMTL under these alternative assumptions is that the auxiliary target $\b{y}'$ is unobserved, as illustrated in Fig.~\ref{fig:backdoor path}. This is problematic because it opens up the backdoor path $\b{y} \leftarrow \b{u} \rightarrow \b{y}' \rightarrow \b{x}$, and makes $\b{x}$ and $\b{y}$ spuriously dependent. Since the backdoor path passes through $\b{u}$, DMTL is sensitive to shifts in $p(\b{u})$. When $p(\b{u})$ shifts, the relationship between $\b{x}$ and $\b{y}$ changes, and DMTL fails to generalize. This is the weakness of DMTL that we aim to improve. Since $\b{u}$ causes the targets, a shift in $p(\b{u})$ leads to a shift in $p(\b{y}, \b{y}')$, which is often called target shift~\citep{Zhang2013ImportanceReweighting}. The targets are observable, unlike $\b{u}$, so we use target shift as an observable proxy for a shift in $p(\b{u})$.

We propose \emph{generative multitask learning} (GMTL) as a way to mitigate target-causing confounding. GMTL is based on the following idea. If we write the causal factorization of the joint distribution of the observed variables as
\begin{align*}
    p(\b{x}, \b{y}, \b{y}') = p(\b{x} \mid \b{y}, \b{y}') p(\b{y}, \b{y}') = p(\b{x} \mid \b{y}, \b{y}') \int p(\b{y} \mid \b{u}) p(\b{y}' \mid \b{u}) p(\b{u}) \mathrm{d}\b{u},
\end{align*}
this shows that $p(\b{x} \mid \b{y}, \b{y}')$ is invariant to shifts in $p(\b{u})$. Therefore, if we use
\begin{align}
    \label{eq:gmtl original}
    \argmax_{\b{y}, \b{y}'} \log p(\b{x} \mid \b{y}, \b{y}')
\end{align}
as the inference objective, we become invariant to shifts in $p(\b{u})$. Importantly, unlike Eq.~\ref{eq:dmtl factorized}, the inference objective in Eq.~\ref{eq:gmtl original} no longer factorizes over the targets. This means that even if we only care about evaluating the main target $\b{y}$, we also need to consider $\b{y}'$. Conditioning on all targets during inference closes the backdoor paths between the inputs and targets. This $d$-separates $\b{x}$ and $\b{u}$, and removes the spurious dependencies between the input and targets.

In practice, we implement GMTL as an approximation of Eq.~\ref{eq:gmtl original}. As we discuss in Section~\ref{sec:GMTL}, this allows us to formulate GMTL as a minor change to the inference objective of DMTL. This makes GMTL practical, since it can be used with existing MTL methods without requiring additional training. We also introduce a single hyperparameter to adjust the degree to which we remove spurious dependencies between the input and targets. This makes GMTL include DMTL as a special case, where no spurious dependencies are removed. This flexibility is useful for managing the trade-off between in-distribution (ID) and out-of-distribution (OOD) generalization, since spurious dependencies are predictive in the former setting, but not the latter.

In order to empirically validate GMTL, we perform experiments on two datasets called Attributes of People~\citep{Bourdev2011AttributesOfPeople} and Taskonomy~\citep{Zamir2018Taskonomy}. Our results show that GMTL improves robustness to target shift across four MTL methods. We attribute this improvement to GMTL's ability to mitigate target-causing confounding. This serves as evidence that causal representation learning shows promise for MTL.

\section{Background}

\subsection{Out-of-distribution generalization}
\label{sec:ood generalization}

OOD generalization refers to the capability of ML systems to generalize on data outside of the training distribution. \citet{Bengio2021DLForAI} describe it as one of the greatest challenges in ML, and one that cannot be solved solely by scaling up datasets and computation. The failure of ML systems to generalize OOD is a general problem that has been reported across a wide range of problems~\citep{DAmour2020Underspecification}. The core issue is that it is often easier for ML systems to learn patterns that are dataset-specific, rather than those that hold universally, and there is often no incentive to learn the latter. These dataset-specific patterns have many names, such as surface statistical regularities~\citep{Jo2017SurfaceStatisticalRegularities}, nonrobust features~\citep{Tsipras2019RobustnessTradeoff}, and shortcuts~\citep{Geirhos2020ShortcutLearning}. We take a causal perspective and adopt the term \emph{spurious dependencies}, which are defined as statistical dependencies not supported by causal links~\citep{Pearl2009Causality}.

\subsection{Dataset biases induce spurious dependencies}
\label{sec:dataset biases}

Spurious dependencies can arise from design choices in dataset creation. \citet{Recht2019ImageNetGeneralize} demonstrated that state-of-the-art image classifiers trained on ImageNet~\citep{Deng2009ImageNet} failed to generalize to a replicated test set designed to mirror the original test set distribution. \citet{Engstrom2021DatasetReplicationBias} offered an explanation for this phenomenon. The authors reported that the dataset replication procedure relied on estimating a human-in-the-loop metric called selection frequency, and that bias in estimating this statistic led to undesirable variation across datasets. Selection frequency is defined for an input-target pair, and measures the proportion of crowdsourced annotators who deem the pair correctly labeled. \citet{Recht2019ImageNetGeneralize} performed statistic matching on this metric, which corresponds to conditioning on it. Since selection frequency is caused by the input and target, this corresponds to conditioning on a collider, which induces a spurious dependency between the input and target. Such a dependency would fail to generalize OOD, which may explain why predictive performance degraded on the replicated test set.

In this work, we address a different source of dataset bias that we call target-causing confounding. A variable is a target-causing confounder if it is causes the targets, but not the input. For a simple example, suppose we have images of people in indoor scenes, and the targets are clothing-related attributes such as hat and scarf. The season causes the targets, since people pair various types of clothing depending on the season. The season does not cause the image, since the images are of indoor scenes. Therefore, the season is a target-causing confounder that induces spurious dependencies between the input and targets. Relying on such dependencies for prediction can be problematic if the training and test sets represent different seasons. Although this is a contrived example for the purpose of illustration, dataset biases can occur in real-life settings like medicine, and have serious societal impacts~\citep{Mehrabi2021BiasFairness}.

\subsection{Causal representation learning}

In Section~\ref{sec:ood generalization}, we made the distinction between patterns that are dataset-specific, and those that hold universally. The field of causality formalizes the notion of patterns that hold universally, calling them causal relations~\citep{Pearl2009Causality}. Causal representation learning aims to use ideas from causality to improve machine learning methods, particularly in the area of OOD generalization~\citep{Zhang2013ImportanceReweighting,Scholkopf2019CausalityML,Scholkopf2021CRL,Arjovsky2019IRM,Puli2021NuisanceSpuriousCorrelations}. Our work falls under this category, since GMTL improves robustness to target shift by using an intervention distribution for prediction. This idea is also shared by \citet{Subbaswamy2019Transport}, who propose a general framework for estimating intervention distributions for robust prediction. GMTL is a practical method for achieving this on high-dimensional datasets. Another causality-motivated paper for robust prediction is \citet{Makar2022ShortcutRemoval}, who use importance reweighting to learn distributions that are invariant to intervention on an auxiliary target. Both \citet{Subbaswamy2019Transport} and \citet{Makar2022ShortcutRemoval} assume the same causal graph as GMTL.

\subsection{Multitask learning}

MTL can reduce computational cost through parameter sharing, and improve predictive performance over training on each task individually~\citep{Standley2020WhichTasksMTL}. Its use is widespread, spanning domains such as scene understanding~\citep{Chowdhuri2019MultiNet,Zamir2018Taskonomy,Zamir2020CrossTaskConsistency}, medical diagnosis~\citep{Kyono2021MTLMammography}, natural language processing~\citep{McCann2018Decathlon,Wang2019GLUE,Radford2019GPT2,Liu2019MTDNN,Worsham2020NLPMTL,Aghajanyan2021MUPPET}, recommender systems~\citep{Covington2016MTLRecSys,Ma2018MGMOE}, and reinforcement learning~\citep{Kalashnikov2021MTOpt}. Distinguishing between main and auxiliary targets is common~\citep{Ruder2017MTLOverview,Liebel2018AuxiliaryTasks,Vafaeikia2020AuxiliaryTaskReview}, but not universal, and is primarily important for concerns such as which loss to use for early stopping~\citep{Caruana1997MTL}. We make this distinction throughout this paper, denoting the main target as $\b{y}$, and the auxiliary target as $\b{y}'$.

\section{Generative multitask learning}
\label{sec:GMTL}

In GMTL, we condition on all targets in order to $d$-separate $\b{x}$ and $\b{u}$, and remove spurious dependencies between the input and targets. During inference, we rely solely on $p(\b{x} \mid \b{y}, \b{y}')$, since is invariant to shifts in $p(\b{u})$. Approaching this naively by estimating $p(\b{x} \mid \b{y}, \b{y}')$ directly can be difficult, especially when $\b{x}$ is high-dimensional. We can get around this by using Bayes' rule to write
\begin{align}
    \label{eq:gmtl bayes}
    \argmax_{\b{y}, \b{y}'} \log p(\b{x} \mid \b{y}, \b{y}') &= \argmax_{\b{y}, \b{y}'} \log \frac{p(\b{y}, \b{y}' \mid \b{x}) p(\b{x})}{p(\b{y}, \b{y}')}\nonumber\\
    &= \argmax_{\b{y}, \b{y}'} \log p(\b{y}, \b{y}' \mid \b{x}) - \log p(\b{y}, \b{y}').
\end{align}
This shows that the argmax over the targets results in the difficult-to-estimate $p(\b{x})$ term to drop out, meaning we only need to estimate $p(\b{y}, \b{y}' \mid \b{x})$ and $p(\b{y}, \b{y}')$.

In order to make GMTL practical, we replace Eq.~\ref{eq:gmtl bayes} with an approximation, i.e.
\begin{align}
    \label{eq:gmtl final}
    \argmax_{\b{y}, \b{y}'} \log p(\b{y} \mid \b{x}) + \log p(\b{y}' \mid \b{x}) - \alpha \log p(\b{y}, \b{y}'), \qquad \alpha \in [0, 1].
\end{align}
This introduces two changes. The first is that we allow $p(\b{y}, \b{y}' \mid \b{x})$ to factorize, as was the case in DMTL. This allows us to use conventional approaches to estimate $p(\b{y}, \b{y}' \mid \b{x})$, which is beneficial since training is notoriously difficult in MTL~\citep{Standley2020WhichTasksMTL}. To be clear, we do not allow this factorization because we assume the targets are conditionally independent given the input, as this would violate our causal graph in Fig.~\ref{fig:gmtl}. We do so purely out of convenience, so that GMTL can leverage existing approaches for estimating $p(\b{y}, \b{y}' \mid \b{x})$. We can allow this factorization as long as we capture the dependency between the targets when estimating $p(\b{y}, \b{y}')$. This prevents Eq.~\ref{eq:gmtl final} from factorizing over the targets, and satisfies our requirement of conditioning on all targets in order to remove spurious dependencies between the input and targets. We discuss some complications with estimating $p(\b{y}, \b{y}')$ in Section~\ref{sec:estimating target distribution}, but in general, it can be done trivially for categorical targets. Therefore, GMTL only changes the inference objective of DMTL, and can be applied with existing multitask methods without requiring additional training.

The second change is the introduction of the hyperparameter $\alpha \in [0, 1]$, which enables us to interpolate between DMTL $(\alpha = 0)$ and the most extreme case of GMTL $(\alpha = 1)$. $\alpha$ controls the degree to which we remove spurious dependencies between the input and targets. The ability to adjust this effect is important because if the training and test distributions are identical, then we should be willing to use any dependencies available, regardless of whether they are spurious or causal. However, it is more realistic to assume training and test distributions will be different, and under these circumstances, setting $\alpha > 0$ can improve robustness to target shift.

\subsection{Interpreting $\alpha$}
\label{sec:interpreting alpha}

The hyperparameter $\alpha$ represents our uncertainty about the test target distribution. In the case of complete certainty, that is, if we know that the test target distribution is identical to the empirical target distribution, we set $\alpha = 0$ to use the empirical distribution. In contrast, in the case of complete uncertainty, we set $\alpha = 1$ to completely remove the influence of the empirical distribution. This is equivalent to assuming the test target distribution is uniform. Generally, it can be shown that GMTL corresponds to replacing the empirical target distribution $p(\b{y}, \b{y}')$ with a distribution proportional to $p(\b{y}, \b{y}')^{1 - \alpha}$. The derivation is in the supplementary material. We perform this replacement during inference in order to keep training the same as DMTL. In principle, we could achieve the same invariance by reweighting examples by $1 / p(\b{y}, \b{y}')^\alpha$ during training~\citep{Zhang2013ImportanceReweighting,Makar2022ShortcutRemoval}.

This raises an important point - $\alpha$ represents an assumption, and is not a hyperparameter that can be tuned in the conventional sense. Instead, tuning $\alpha$ is analogous to model selection w.r.t. an unknown distribution, which is a difficult open problem~\citep{Gulrajani2021LostDG}. Progress in this direction is likely to benefit GMTL, as well as other methods with similar hyperparameters~\citep{Wortsman2022FineTuningCLIP}. In the remainder of this work, we assume oracle access to the optimal $\alpha$ for a given OOD test distribution. That is, our main focus is not on the difficulty of tuning $\alpha$, which is a separate research question. Instead, we study the efficacy of GMTL assuming we have correctly assessed how similar the training and test target distributions are, i.e. knowing the optimal $\alpha$.

Having said that, we provide a simple and effective heuristic for setting $\alpha$ even for an unknown test distribution, which makes GMTL practical. This is based on the fact that $\alpha$ represents a trade-off. Increasing $\alpha$ improves target shift robustness at the expense of ID predictive performance. We show empirically in Section~\ref{sec:results} that this relationship is very strong, and holds across two datasets, multiple pairs of tasks, and four MTL methods. Therefore, our heuristic is to set $\alpha$ to the maximum value within an acceptable loss in ID predictive performance, where the latter is measured using the validation set. This ensures that we are as robust as possible to target shift, while keeping ID predictive performance at an acceptable level. These results are in the supplementary material.

If we do have access to the test target distribution $q(\b{y}, \b{y}')$, then we can do away with $\alpha$ and simply replace $p(\b{y}, \b{y}')$ with $q(\b{y}, \b{y}')$, which results in the following inference objective:
\begin{align*}
    \argmax_{\b{y}, \b{y}'} \log p(\b{y} \mid \b{x}) + \log p(\b{y}' \mid \b{x}) - \log p(\b{y}, \b{y}') + \log q(\b{y}, \b{y}').
\end{align*}
Alternatively, we can use $q(\b{y}, \b{y}') / p(\b{y}, \b{y}')$ to correct for target shift during training~\citep{Zhang2013ImportanceReweighting}.

\subsection{Estimating $p(\b{y}, \b{y}')$}
\label{sec:estimating target distribution}

GMTL requires estimating $p(\b{y}, \b{y}')$, which is trivial for categorical targets when $\Y{\times}\Y'$ is small. The maximum likelihood estimate of $p(\b{y}, \b{y}')$ is obtained by counting the number of occurrences of the pair $(\b{y}, \b{y}')$, and dividing by the total across all pairs. However, there is an important precaution to be made, since it is possible that certain pairs of $(\b{y}, \b{y}')$ are never observed. This is problematic, since $p(\b{y}, \b{y}') = 0$ results in taking a logarithm of zero in Eq.~\ref{eq:gmtl final}. This can be addressed with additive smoothing, where a pseudocount $\epsilon > 0$ is added to the count for each pair prior to normalization. The issue of sparsity in the target distribution is the primary challenge in scaling GMTL to a larger number of tasks. Another complication is that in MTL, it is not unusual for a subset of targets to be missing for a given example. When this is the case, the counting approach only works with the subset of examples for which all targets are present, which can lead to poor estimates.

\section{Illustrative example}

Before empirically validating GMTL on real datasets, we use synthetic data to demonstrate how DMTL and GMTL behave differently in the presence of target-causing confounding. We construct a data generating process (DGP) where a target-causing confounder induces a spurious dependency between a scalar input $X$ and binary target $Y$.

Suppose there are two targets $Y, Y' \in \{0, 1\}$ that are bivariate Bernoulli. This distribution can be specified in terms of $P(Y = 1) = \theta$, $P(Y' = 1) = \theta'$, and $\Cov[Y, Y']$. We keep $\theta$ and $\theta'$ fixed, and vary the value of $\Cov[Y, Y']$ in order to induce target shift. $\Cov[Y, Y']$ is a target-causing confounder in this context. $X$ is generated by a mixture of Gaussians with latent variable $Z = 2Y + Y'$, i.e. $X \mid z \sim N(z, \sigma_z^2)$. We fix $\theta = \theta' = 0.5$, $\sigma_0^2 = \sigma_1^2 = 0.4$, and $\sigma_2^2 = \sigma_3^2 = 0.6$, and vary $\Cov[Y, Y']$ from $-0.2$ to $0.2$. In other words, the only property of the DGP that we vary is $\Cov[Y, Y']$.

\begin{figure}[h]
    \centering
    \includegraphics[width=0.5\textwidth]{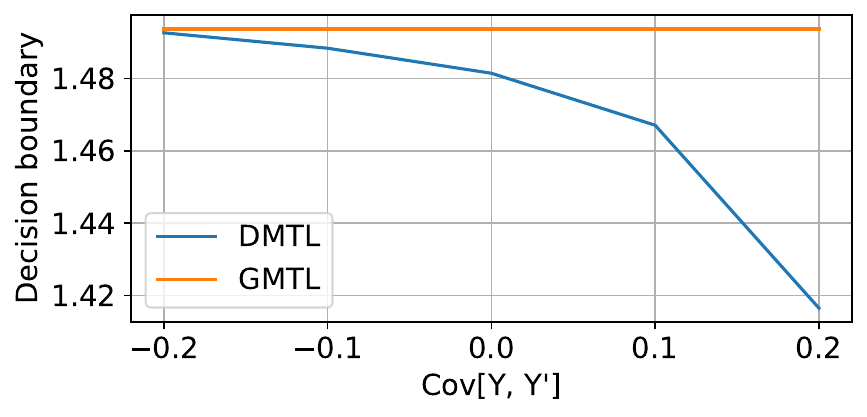}
    \caption{Here we show how the decision boundaries of GMTL and DMTL change w.r.t. $\Cov[Y, Y']$.  DMTL is sensitive to changes in $\Cov[Y, Y']$, while GMTL is invariant.}
    \label{fig:illustrative_example}
\end{figure}

Fig.~\ref{fig:illustrative_example} shows how the decision boundaries of DMTL and GMTL with $\alpha = 1$ are affected by variation in the target-causing confounder $\Cov[Y, Y']$. The decision boundary of DMTL is the value of $X$ for which the solution to $\argmax_y \log P(y \mid x)$ changes, where $Y'$ has been marginalized out. In contrast, the decision boundary of GMTL is the value of $X$ where the $Y$ component of the solution to $\argmax_{y, y'} \log P(x \mid y, y')$ changes. The decision boundary for DMTL changes in response to $\Cov[Y, Y']$, while it remains constant for GMTL. This is because GMTL only uses the Gaussian densities for prediction, which have no connection to $\Cov[Y, Y']$.

In this example, we have shown that GMTL and DMTL arrive at different solutions by varying only the dependency between $Y$ and $Y'$ induced by the target-causing confounder $\Cov[Y, Y']$. The key difference between the two methods is that DMTL uses the spurious dependency between $X$ and $Y$, while GMTL does not. Which of the two methods achieves better predictive performance depends on the particular test distribution, since the spurious dependency will be predictive in ID settings, and not in OOD settings.

\section{Experimental setup}
\label{sec:experimental setup}

We perform experiments to demonstrate that GMTL improves robustness to target shift. We evaluate across a wide range of target shifts ranging from mild to severe, and analyze how the optimal $\alpha$ changes w.r.t. the severity of target shift. Recall from Section~\ref{sec:interpreting alpha} that we assume oracle access to the optimal $\alpha$, since we are focusing on validating GMTL, not on whether we can perform model selection w.r.t. unknown distributions, which is a separate open problem. The purpose of these experiments is to show that if we correctly assess how similar the training and test distributions are, we can benefit from using GMTL. In order to draw robust conclusions, we perform our experiments using two datasets, multiple pairs of classification tasks, and four MTL methods. Here, we describe our experimental setup.

\subsection{Datasets and tasks}

\paragraph{Attributes of People} The Attributes of People dataset~\citep{Bourdev2011AttributesOfPeople} consists of 8,035 images of people. There are 4,013 examples in the training set, and 4,022 in the test set. Since the authors did not specify a validation set, we randomly sample $20\%$ of the training set to use as the validation set. Each image comes labeled with up to nine binary attributes: male, long hair, glasses, hat, t-shirt, long sleeves, shorts, jeans, and long pants. We use each attribute as a binary classification task, and predict whether the attribute appears in the image. We drop the gender attribute, since it has the potential for negative downstream applications~\citep{Wang2018SexualOrientation}. We experiment with various pairs of tasks, taking turns specifying one as the main task, and the other as the auxiliary task. For each pair of tasks, we report the classification accuracy for the main task. Since not all attributes are labeled for each example, we consider the three pairs of tasks with the highest number of co-occurrences. These are: hat and long sleeves, long hair and hat, and glasses and hat.

\paragraph{Taskonomy} The Taskonomy dataset~\citep{Zamir2018Taskonomy} is of a much larger scale than the first dataset, containing approximately four million images of indoor scenes. We use a subset of the dataset that is provided by the authors for faster experimentation, with 548,785 examples in the training set, 121,974 in the validation set, and 90,658 in the test set. Each image comes paired with $26$ labels relevant to scene understanding, but since we are focusing on classification, we only use the object and scene annotations as targets. We use the object annotations as a $100$-way object classification task, and the scene annotations as a $64$-way scene classification task. Between object and scene classification, we take turns specifying one as the main task, and the other as the auxiliary task. We report the top-$1$ accuracy for the main task.

\subsection{Multitask learning methods}

\paragraph{No Parameter Sharing (NPS)} The first MTL method we use is a trivial combination of two single-task networks. That is, $p(\b{y} \mid \b{x})$ and $p(\b{y}' \mid \b{x})$ are trained separately. We call this No Parameter Sharing (NPS). This is equivalent to single-task learning during training, but becomes different during inference for $\alpha > 0$, since the single-task networks influence one another through the $-\alpha \log p(\b{y}, \b{y}')$ term. While NPS is primarily meant to be a simple setting for comparing GMTL and DMTL, it offers some practical utility as well. MTL methods are notoriously difficult to train, and it is often challenging to get them to perform better than the single-task baseline~\citep{Alonso2017WhenMTLEffective}. NPS combined with GMTL offers a way to combine two single-task networks to reap the benefits of MTL, just by estimating $p(\b{y}, \b{y}')$ and predicting jointly over the targets.

\paragraph{Shared Trunk Networks (STN)} The second MTL method is a hard parameter sharing method that we call Shared Trunk Networks (STN), where all weights except for the final classification layer are shared. This is not an effective architecture for many of the tasks, since the predictive performance is worse than the single-task baseline. This is common for STN~\citep{Alonso2017WhenMTLEffective}, but we include it for completeness because it is the canonical MTL architecture. The results comparing single-task and MTL methods on the ID test set are in the supplementary material.

\paragraph{Cross-stitch Networks (CSN)} For the third MTL method, we use a soft parameter sharing method called Cross-stitch Networks (CSN)~\citep{Misra2016CrossStitchNetworks}. CSN takes two trained single-task networks, and takes a linear combination of the activations at each layer during the forward pass. Training CSN involves learning the coefficients of the linear combinations, as well as fine-tuning the single-task network weights. This network performs better than the single-task baseline across most datasets and tasks.

\paragraph{Full Parameter Sharing (FPS)} For our experiments on Attributes of People, we model $p(\b{y}, \b{y}' \mid \b{x})$ directly as a four-class classification, without assuming factorization. Since this is a single network with all weights shared across both tasks, we call this Full Parameter Sharing (FPS). We include FPS to assess the appropriateness of letting $p(\b{y}, \b{y}' \mid \b{x})$ factorize in the other MTL approaches, as we discussed in Section~\ref{sec:GMTL}. We omit FPS on Taskonomy due to the complications of $\Y{\times}\Y'$ being large, and leave this to future work.

\subsection{Methodology for training and prediction}

For all datasets and tasks, we use ResNet-50~\citep{He2016ResNet} pretrained on ImageNet~\citep{Deng2009ImageNet} for the single-task networks. We train using Adam~\citep{Kingma2015Adam} with $L_2$ regularization. We tune the learning rate and regularization multiplier. For data augmentation, during training we resize the images to $256{\times}256$, randomly crop them to $224{\times}224$ and randomly horizontally flip them. During validation and testing, we resize the images to $256{\times}256$, and center crop them to $224{\times}224$. For all experiments, we train single-task networks from five random initializations. Details regarding other hyperparameters are included in the supplementary material.

\subsection{Simulating target shift with importance sampling}

In our experiments, we aim to evaluate across a wide range of target shifts. Since it is impractical to collect many different test sets, we simulate target shifts using importance sampling. Suppose the original test distribution is $p(\b{x}, \b{y}, \b{y}') = p(\b{x} \mid \b{y}, \b{y}') p(\b{y}, \b{y}')$, and the simulated test distribution is $q(\b{x}, \b{y}, \b{y}') = p(\b{x} \mid \b{y}, \b{y}') q(\b{y}, \b{y}')$. $p(\b{x} \mid \b{y}, \b{y}')$ is invariant due to our assumption that it is a causal relation. Therefore, the expected accuracy $l(\b{x}, \b{y}, \b{y}')$ under the simulated test distribution $q(\b{x}, \b{y}, \b{y}')$ is given by
\begin{align*}
    \E_{q(\b{x}, \b{y}, \b{y}')}[l(\b{x}, \b{y}, \b{y}')] &= \E_{p(\b{x} \mid \b{y}, \b{y}') q(\b{y}, \b{y}')}[l(\b{x}, \b{y}, \b{y}')]\\
    &= \E_{p(\b{x}, \b{y}, \b{y}')}\left[\frac{q(\b{y}, \b{y}')}{p(\b{y}, \b{y}')} l(\b{x}, \b{y}, \b{y}')\right]\\
    &\approx \frac{1}{N} \sum_{n=1}^N \frac{q(\b{y}^{(n)}, \b{y}'^{(n)})}{p(\b{y}^{(n)}, \b{y}'^{(n)})} l(\b{x}^{(n)}, \b{y}^{(n)}, \b{y}'^{(n)})\\
    &\approx \sum_{n=1}^N \frac{q(\b{y}^{(n)}, \b{y}'^{(n)})}{p(\b{y}^{(n)}, \b{y}'^{(n)})} \bigg/ \sum_{m = 1}^N \frac{q(\b{y}^{(m)}, \b{y}'^{(m)})}{p(\b{y}^{(m)}, \b{y}'^{(m)})} l(\b{x}^{(n)}, \b{y}^{(n)}, \b{y}'^{(n)}).
\end{align*}
Now that we can evaluate across a wide range of OOD $q(\b{y}, \b{y}')$'s, we need a way to quantify the severity of the target shift between $p(\b{y}, \b{y}')$ and $q(\b{y}, \b{y}')$.

\subsection{Measuring the severity of target shift between $p(\b{y}, \b{y}')$ and $q(\b{y}, \b{y}')$}

Our metric to measure the severity of target shift between $p(\b{y}, \b{y}')$ and $q(\b{y}, \b{y}')$ must be such that predictive performance degrades w.r.t. the severity of shift. Not all metrics satisfy this criteria. To build intuition, consider a Bernoulli distribution where $P(Y = 1) = \theta$. If we use a norm-based metric such as total variation, this treats a change in $\theta$ from 0.6 to 0.8 and a change from 0.6 to 0.4 as being the same. However, from the point of view of prediction, the latter should be more detrimental, since it reverses the ranking of the classes.

Similarly, Kullback-Leibler (KL) divergence is a common choice for measuring the distance between distributions, but it suffers from a related problem. Consider $\theta$ from above changing from $0.9$ to $0.9999$, and from $0.9$ to $0.4$. Due to taking the logarithm of a small number, KL divergence considers the first to be a more severe shift. This is undesirable, because the change from $0.9$ to $0.4$ represents a reversal in the ranking of the classes.

These examples point to a need for a ranking-based metric, since a significant change to the ranking of classes is detrimental to predictive performance. Weighted rank correlation satisfies this. If the weighted rank correlation between $p(\b{y}, \b{y}')$ and $q(\b{y}, \b{y}')$ is positive with large magnitude, it means that there are no significant changes in the ranking of classes, and that the target shift is not severe. In contrast, if the weighted rank correlation is negative with large magnitude, it implies the rankings of classes has changed significantly. This constitutes a more severe target shift. We therefore use weighted Kendall's $\tau$~\citep{Shieh1998WeightedTau}, which we henceforth refer to as $\tau$, to measure the severity of target shift between $p(\b{y}, \b{y}')$ and $q(\b{y}, \b{y}')$. We provide the definition of $\tau$ in the supplementary material. It relies on a weighting function, for which we use $1 / (r + 1)$, where $r$ is the ranking.

\subsection{Sampling $q(\b{y}, \b{y}')$}

Our choice of distance metric informs how to sample $q(\b{y}, \b{y}')$, since our goal is to evaluate across a wide range of target shift severities. We designed a method which shuffles $\log p(\b{y}, \b{y}')$ and perturbs it with noise. To be precise, let $f(\b{y}, \b{y}') \in \{1, 2, \dotsc, \lvert \Y \times \Y' \rvert\}$ such that each $(\b{y}, \b{y}')$ is assigned a unique integer index. We use $p_{f(\b{y}, \b{y}')} = p(\b{y}, \b{y}')$ Let $i = \lambda \cdot \lvert \Y \times \Y' \rvert$, where $\lambda \sim \U(0, 0.5)$. We randomly shuffle these indices with $\pi: \{1, 2, \dotsc, i\} \to \{1, 2, \dotsc, \lambda \times i\}$. Then, we define $q(\b{y}, \b{y}')$ by
\begin{align*}
q(\b{y}, \b{y}')
\propto
\begin{cases}
\exp(\log p(\b{y}, \b{y}') + \epsilon) & \text{if} \ f(\b{y}, \b{y}') > i \\
\exp(\log p_{\pi(f(\b{y}, \b{y}'))} + \epsilon) & \text{if} \ f(\b{y}, \b{y}') \leq i,
\end{cases}
\end{align*}
where $\epsilon \sim \N(0, \sigma^2)$ and $\sigma \sim \U(10^{-12}, 5)$.

\section{Results}
\label{sec:results}

\begin{figure}[t]
    \centering
    \includegraphics[width=\textwidth]{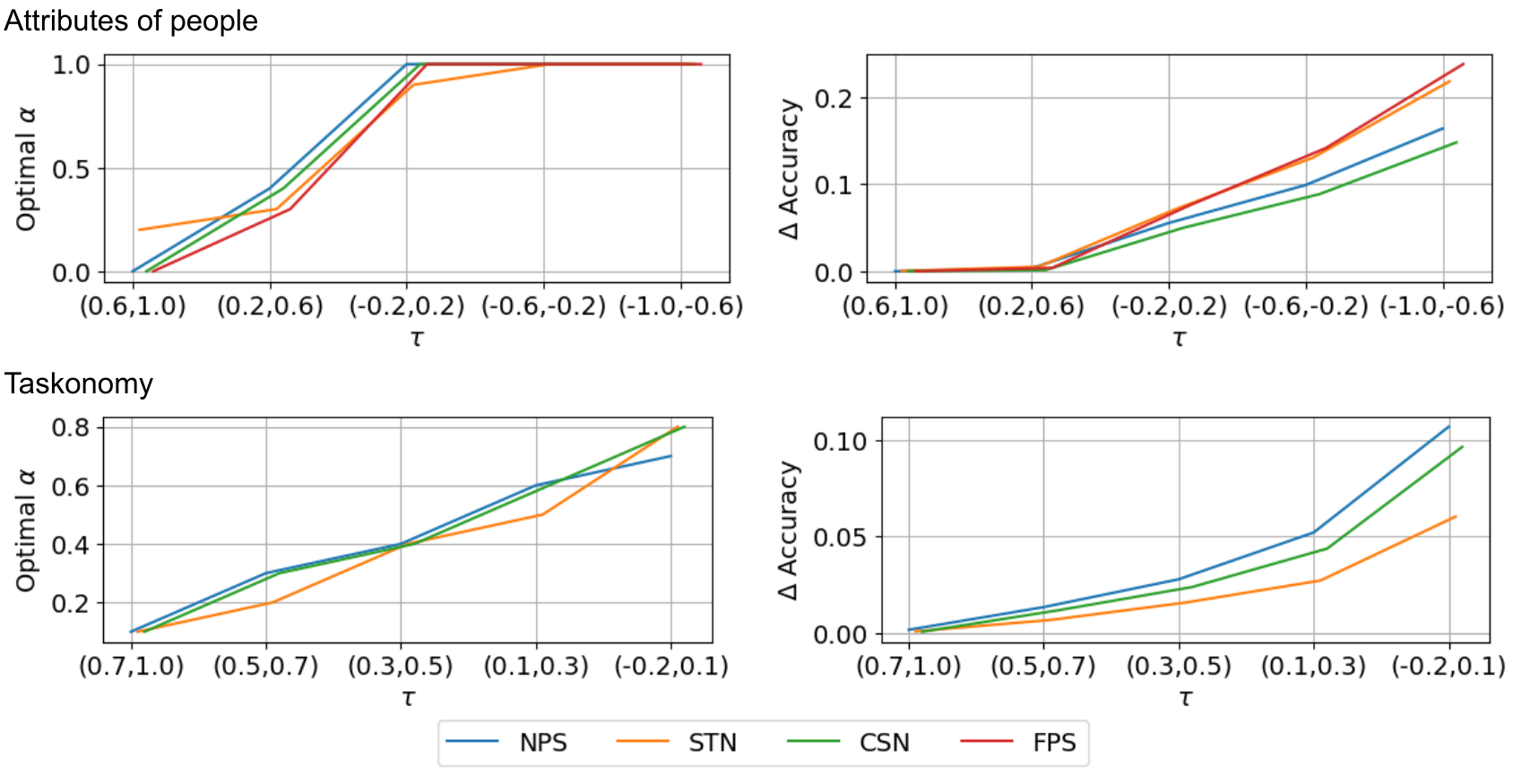}
    \caption{In order to evaluate whether GMTL improves robustness to target shift, we estimate OOD accuracy across a range of target shift severities and a range of $\alpha \in [0, 1]$. The results are grouped w.r.t. the severity of target shift $\tau$, and we report the optimal $\alpha$ for each group. As seen in the left subfigures, as the target shift becomes more severe from left to right, the optimal $\alpha$ monotonically increases for both Attributes of People (top left) and Taskonomy (bottom left). This is because a larger $\alpha$ removes more spurious dependencies induced by target-causing confounding, which is more beneficial when the severity of shift increases. In the right subfigures, the vertical axis is the difference in accuracy between GMTL and DMTL. The gain in accuracy from using GMTL increases w.r.t. the severity of target shift for both Attributes of People (top right) and Taskonomy (bottom right). The reason is that the spurious dependencies induced by target-causing confounding become less predictive as the severity of shift increases, and therefore removing them yields larger improvements in accuracy.}
    \label{fig:optimal_alpha}
\end{figure}

We compute the OOD test set accuracy across target shifts ranging from mild to severe, where $p(\b{y}, \b{y}')$ is the original test set distribution, and $q(\b{y}, \b{y}')$ is a randomly sampled OOD distribution. First, we sample an OOD $q(\b{y}, \b{y}')$, and compute $\tau$ relative to $p(\b{y}, \b{y}')$. $\tau$ represents the severity of target shift between $q(\b{y}, \b{y}')$ and the original test distribution $p(\b{y}, \b{y}')$. We then use importance sampling to compute the OOD accuracy under each $q(\b{y}, \b{y}')$ across a range of $\alpha \in [0, 1]$.

The results are split into five groups that are equally spaced in terms of $\tau$. Each group represents the accuracies for a range of $\alpha$ under a particular severity of target shift. For each group, we compute the optimal value of $\alpha$ that attains the highest accuracy. We also compute the difference in accuracy between GMTL and DMTL at the optimal value of $\alpha$ for each group.

We observe strikingly similar results for both datasets and all pairs of tasks. As the target shift becomes more severe, the optimal $\alpha$ increases. A representative example for each dataset is shown in the left subfigures in Fig.~\ref{fig:optimal_alpha}, and the rest are in the supplementary material. This observation matches our understanding of $\alpha$ from Section~\ref{sec:interpreting alpha} - since $\alpha$ is our degree of belief on how similar $p(\b{y}, \b{y}')$ and $q(\b{y}, \b{y}')$ are, the optimal $\alpha$ tends to be higher when the target shift is more severe. Increasing $\alpha$ removes more of the spurious dependencies induced by target-causing confounding, which is more beneficial for severe target shifts.

Also, as the severity of target shift increases, the difference in accuracy between GMTL and DMTL for the optimal $\alpha$ increases as well. This can be seen in the right subfigures in Fig.~\ref{fig:optimal_alpha}, and is also in-line with our expectations. As the target shift becomes more severe, the spurious dependencies induced by target-causing confounding become less predictive, and removing them yields larger gains in predictive performance.

Both patterns hold very similarly for all MTL methods. This suggests that it is our formulation of GMTL that is significant, rather than the particular parameterization used to learn $p(\b{y}, \b{y}' \mid \b{x})$. Both the optimal $\alpha$, as well as the corresponding gain in accuracy using GMTL, increase monotonically w.r.t. the severity of target shift. This is a strong relation that holds across datasets, pairs of tasks, and MTL methods. To quantify the generality of this result, we aggregate all of our results, and compute the correlation between $\tau$ and the optimal $\alpha$. Since $\tau$ is used in the context of an interval, we take the midpoint of each interval. This results in a correlation of $-0.847$. This indicates a strong positive correlation between the severity of target shift and the optimal $\alpha$. The generality of this relation is strong evidence that GMTL improves robustness to target-causing confounding, and also that our interpretation of $\alpha$ is correct. Our experimental results using the $\alpha$-selection heuristic described in Section~\ref{sec:interpreting alpha}, as well as other results showing the accuracy for a range of $\alpha$, not just the optimal one, are in the supplementary material.

\section{Conclusion}

We presented generative multitask learning (GMTL), an approach for causal representation learning for MTL that only changes the inference objective of conventional MTL. Our approach mitigates the effect of target-causing confounders, which are variables that cause the targets, but not the input. This removes spurious dependencies between the input and targets, and improves robustness to target shift. Looking forward, we are excited by the new perspectives that causal representation learning can bring to improve weaknesses in ML. Our work demonstrates the potential for this in the context of MTL. We plan to investigate what other weaknesses in ML can be improved by adopting a causal perspective.

\section{Acknowledgements}

This work was supported by grants from the National Institutes of Health (P41EB017183), the National Science Foundation (HDR-1922658), the Gordon and Betty Moore Foundation (9683), and Samsung Advanced Institute of Technology (Next Generation Deep Learning: From Pattern Recognition to AI).

\bibliography{main}
\bibliographystyle{plainnat}

\begin{enumerate}

\item For all authors...
\begin{enumerate}
\item Do the main claims made in the abstract and introduction accurately reflect the paper's contributions and scope?
    \answerYes{}
\item Did you describe the limitations of your work?
    \answerYes{See Section~\ref{sec:interpreting alpha}}
\item Did you discuss any potential negative societal impacts of your work?
    \answerYes{See Section~\ref{sec:dataset biases}}
\item Have you read the ethics review guidelines and ensured that your paper conforms to them?
    \answerYes{}
\end{enumerate}

\item If you are including theoretical results...
\begin{enumerate}
\item Did you state the full set of assumptions of all theoretical results?
    \answerNA{}
\item Did you include complete proofs of all theoretical results?
    \answerNA{}
\end{enumerate}

\item If you ran experiments...
\begin{enumerate}
\item Did you include the code, data, and instructions needed to reproduce the main experimental results (either in the supplemental material or as a URL)?
    \answerYes{See supplementary material}
\item Did you specify all the training details (e.g., data splits, hyperparameters, how they were chosen)?
    \answerYes{See Section~\ref{sec:experimental setup} and the supplementary material}
\item Did you report error bars (e.g., with respect to the random seed after running experiments multiple times)?
    \answerYes{Our main results in Section~\ref{sec:results} do not allow error bars, but they represent the combined results of five random seeds. Our results in the supplementary material do have error bars w.r.t. the random seeds.}
\item Did you include the total amount of compute and the type of resources used (e.g., type of GPUs, internal cluster, or cloud provider)?
    \answerYes{See supplementary material}
\end{enumerate}

\item If you are using existing assets (e.g., code, data, models) or curating/releasing new assets...
\begin{enumerate}
\item If your work uses existing assets, did you cite the creators?
    \answerYes{See Section~\ref{sec:experimental setup}, we use two public datasets: Attributes of People and Taskonomy.}
\item Did you mention the license of the assets?
    \answerYes{We mention the license of the two public datasets in the supplementary material.}
\item Did you include any new assets either in the supplemental material or as a URL?
    \answerYes{We include our code in the supplementary material.}
\item Did you discuss whether and how consent was obtained from people whose data you're using/curating?
    \answerNA{No need to, the datasets are public.}
\item Did you discuss whether the data you are using/curating contains personally identifiable information or offensive content?
    \answerYes{See Section~\ref{sec:experimental setup}, the Attributes of People dataset has a gender label that we drop, since predicting gender can have negative consequences.}
\end{enumerate}

\item If you used crowdsourcing or conducted research with human subjects...
\begin{enumerate}
\item Did you include the full text of instructions given to participants and screenshots, if applicable?
    \answerNA{}
\item Did you describe any potential participant risks, with links to Institutional Review Board (IRB) approvals, if applicable?
    \answerNA{}
\item Did you include the estimated hourly wage paid to participants and the total amount spent on participant compensation?
    \answerNA{}
\end{enumerate}

\end{enumerate}

\pagebreak

\appendix

\section{Appendix}

\subsection{Definitions}

Weighted Kendall's $\tau$ is a measure of rank correlation between two vectors $\b{x}$ and $\b{y}$, both with length $N$. Using the authors' notation, let $(i, R_i)$ for $i = 1, \dotsc, N$ be pairs such that $R_i$ is the rank of the element in $\b{y}$ whose corresponding $\b{x}$ value has rank $i$. Let $w(i, j)$ be a bounded and symmetric weight function that maps to $\R$, and denote its value at $(i, j)$ as $w_{ij}$. Weighted Kendall's $\tau$ is defined as
\begin{align*}
    \tau = \frac{\sum_{i \neq j} w_{ij} \sgn(i - j) \sgn(R_i - R_j)}{\sum_{i, j} w_{ij} - \sum_i w_{ii}},
\end{align*}
where
\begin{align*}
    \sgn(x) = 
    \begin{cases}
    -1 & \text{if} \ x < 0\\
    0 & \text{if} \ x = 0\\
    1 & \text{if} \ x > 0.
    \end{cases}
\end{align*}

\subsection{Derivations}

GMTL corresponds to replacing the empirical target distribution $p(\b{y}, \b{y}')$ with a distribution that is proportional to $p(\b{y}, \b{y}')^{1 - \alpha}$. To see why, suppose the empirical joint distribution is $p(\b{x}, \b{y}, \b{y}')$ and we undergo a target shift such that the test distribution is $q(\b{x}, \b{y}, \b{y}') = p(\b{x} \mid \b{y}, \b{y}') q(\b{y}, \b{y}')$. If we assume that $p(\b{y}, \b{y}' \mid x)$ factorizes, then predicting optimally w.r.t. $q(\b{x}, \b{y}, \b{y}')$ gives us
\begin{align}
    \label{eq:importance reweighting}
    \argmax_{\b{y}, \b{y}'} \log q(\b{x}, \b{y}, \b{y}') &= \argmax_{\b{y}, \b{y}'} \log p(\b{x} \mid \b{y}, \b{y}') + \log q(\b{y}, \b{y}')\nonumber\\
    &= \argmax_{\b{y}, \b{y}'} \log p(\b{x} \mid \b{y}, \b{y}') + \log q(\b{y}, \b{y}')\nonumber\\
    &= \argmax_{\b{y}, \b{y}'} \log p(\b{y}, \b{y}' \mid x) - \log p(\b{y}, \b{y}') + \log q(\b{y}, \b{y}')\nonumber\\
    &= \argmax_{\b{y}, \b{y}'} \log p(\b{y} \mid \b{x}) + \log p(\b{y}' \mid \b{x}) \underbrace{- \log p(\b{y}, \b{y}') + \log q(\b{y}, \b{y}')}_{-\alpha \log p(\b{y}, \b{y}')}.
\end{align}
Notice that Eq.~\ref{eq:importance reweighting} resembles GMTL. If we write
\begin{align*}
    -\log p(\b{y}, \b{y}') + \log q(\b{y}, \b{y}') &\propto -\alpha \log p(\b{y}, \b{y}')\\
    \log q(\b{y}, \b{y}') &\propto (1 - \alpha) \log p(\b{y}, \b{y}'),
\end{align*}
this shows that GMTL corresponds to assuming $q(\b{y}, \b{y}') \propto p(\b{y}, \b{y}')^{1 - \alpha}$.

\subsection{Reproducibility}

The code required to fully reproduce our experiments, including the configuration files that contain all hyperparameter values, is available at \url{https://github.com/nyukat/generative-multitask-learning}. We ran our experiments on a single NVIDIA RTX8000 GPU on our high performance computing system. For convenience, here are the three functions relevant to the GMTL inference objective. When executed sequentially, they take as input the task-specific log probabilities $\log p(\b{y} \mid \b{x})$ and $\log p(\b{y}' \mid \b{x})$, the target distribution $p(\b{y}, \b{y}')$, and the parameter $\alpha$, and returns
\begin{align*}
    \argmax_{\b{y}, \b{y}'} \log p(\b{y} \mid \b{x}) + \log p(\b{y}' \mid \b{x}) - \alpha \log p(\b{y}, \b{y}').
\end{align*}

\begin{lstlisting}
def to_log_joint_pred(log_marginals):
    '''
    Input: [log p(y | x), log p(y' | x)]
    Output: log p(y | x) + log p(y' | x)
    '''
    log_joint_shape = [elem.shape[1] for elem in log_marginals]
    n_examples = len(log_marginals[0])
    log_joint = np.full(log_joint_shape + [n_examples], np.nan)
    for flat_idx in range(np.prod(log_joint_shape)):
        unflat_idx = np.unravel_index(flat_idx, log_joint_shape)
        log_prob = 0
        for task_idx, class_idx in enumerate(unflat_idx):
            log_prob += log_marginals[task_idx][:, class_idx]
        log_joint[unflat_idx] = log_prob
    log_joint = np.moveaxis(log_joint, -1, 0)
    return log_joint

def to_generative_pred(log_joint, alpha, log_prior):
    '''
    Input: log p(y | x) + log p(y' | x), alpha, log p(y, y')
    Output: log p(y | x) + log p(y' | x) - alpha * log p(y, y')
    '''
    return log_joint - alpha * log_prior

def to_class_pred(generative_pred):
    '''
    Input: log p(y | x) + log p(y' | x) - alpha * log p(y, y')
    Output: argmax_{y, y'} log p(y | x) + log p(y' | x) - 
        alpha * log p(y, y')
    '''
    class_pred = []
    shape = generative_pred.shape[1:]
    for pred_elem in generative_pred:
        class_pred.append(np.unravel_index(np.argmax(pred_elem), 
            shape))
    class_pred = np.array(class_pred)
    return class_pred
\end{lstlisting}

\subsection{In-distribution test set accuracy}

We report the in-distribution (ID) test set accuracy for each task and MTL method. These results use the test set that originally came with each dataset. The purpose of these results is to show how the MTL methods compare to one another in the ID setting. Since we know that the training and test distributions are similar, we set $\alpha = 0$, in which case NPS is equivalent to single-task learning (STL). STL is a strong baseline, but CSN performs better than it for the majority of tasks. The performance of STN is mixed, which is consistent with the MTL literature. STN does significantly worse than the STL baseline on Taskonomy despite extensive hyperparameter tuning, but we include the results because it is the canonical MTL architecture.

\begin{table}[H]
    \centering
    \caption{Test set accuracy for hat and long sleeves on Attributes of People.}
    \begin{tabular}{lcc}
    \toprule
    & Hat & Long sleeves\\
    \midrule
    NPS & $0.917 \pm 0.003$ & $0.853 \pm 0.004$\\
    STN & $0.912 \pm 0.002$ & $0.852 \pm 0.003$\\
    CSN & $0.920 \pm 0.001$ & $0.861 \pm 0.005$\\
    \bottomrule
    \end{tabular}
    \label{tab:attributes_of_people,t=2,4}
\end{table}

\begin{table}[H]
    \centering
    \caption{Test set accuracy for long hair and hat on Attributes of People.}
    \begin{tabular}{lcc}
    \toprule
    & Long hair & Hat\\
    \midrule
    NPS & $0.858 \pm 0.003$ & $0.917 \pm 0.003$\\
    STN & $0.859 \pm 0.003$ & $0.916 \pm 0.000$\\
    CSN & $0.860 \pm 0.002$ & $0.917 \pm 0.002$\\
    \bottomrule
    \end{tabular}
    \label{tab:attributes_of_people,t=0,2}
\end{table}

\begin{table}[H]
    \centering
    \caption{Test set accuracy for glasses and hat on Attributes of People.}
    \begin{tabular}{lcc}
    \toprule
    & Glasses & Hat\\
    \midrule
    NPS & $0.843 \pm 0.011$ & $0.917 \pm 0.003$\\
    STN & $0.855 \pm 0.003$ & $0.913 \pm 0.002$\\
    CSN & $0.848 \pm 0.001$ & $0.917 \pm 0.002$\\
    \bottomrule
    \end{tabular}
    \label{tab:attributes_of_people,t=1,2}
\end{table}

\begin{table}[H]
    \centering
    \caption{Top-1 test set accuracy for object and scene classification on Taskonomy.}
    \begin{tabular}{lcc}
    \toprule
    & Object & Scene\\
    \midrule
    NPS & $0.749 \pm 0.001$ & $0.730 \pm 0.001$\\
    STN & $0.710 \pm 0.001$ & $0.719 \pm 0.001$\\
    CSN & $0.750 \pm 0.001$ & $0.737 \pm 0.001$\\
    \bottomrule
    \end{tabular}
    \label{tab:taskonomy}
\end{table}

\subsection{Anomalous results}
\label{sec:anomalous results}

As we will see throughout Sections \ref{sec:select alpha}--\ref{sec:alpha range}, the results for Attributes of People with long sleeves as the main task, and hat as the auxiliary task are anomalous. This is because our method for simulating and measuring target shift is not effective for this task. That is, the predictive performance increases w.r.t. the severity of target shift. The problem is that for this task, reversing the roles of the least and most common classes improves predictive performance. This runs counter to our intuition, as well as all other tasks in our experiments. Nonetheless, we include these results for completeness.

\subsection{Results using a heuristic to select $\alpha$}
\label{sec:select alpha}

We report the results from using the $\alpha$-selection heuristic discussed in Section 3.2 of the main text. The heuristic is to choose the maximum possible $\alpha$ within an allowable budget of lost accuracy in the ID setting. For a given accuracy budget, we compute the validation set accuracy for a range of $\alpha$, and pick the largest $\alpha$ such that the accuracy is within the budget relative to $\alpha = 0$. Then, using the selected $\alpha$, we report the test set accuracy across a wide range of target shifts. The horizontal axis in these plots is the severity of target shift, increasing from left to right.

In each set of six figures below, we consider a pair of tasks. One of these tasks is designated the main task, and the other is the auxiliary task. This distinction is primarily important during training, when deciding which task loss to use for early stopping. We then report the accuracy only for the main task (left subfigures). We then reverse the roles of the main and auxiliary tasks for the right subfigures.

For all tasks except for the one mentioned in Section~\ref{sec:anomalous results}, there is a very clear pattern that holds across the three MTL methods. Paying a small penalty in ID accuracy yields significant improvements in robustness to target shift. In some cases the improvement is very large, with roughly a $20\%$ improvement for the most severe target shift (the left subfigures in Fig.~\ref{fig:select_alpha,attributes_of_people,t=0,2}). These results show that even a simple heuristic such as this can be very effective, which makes GMTL practical.

\begin{figure}[H]
    \centering
    \includegraphics[width=\textwidth]{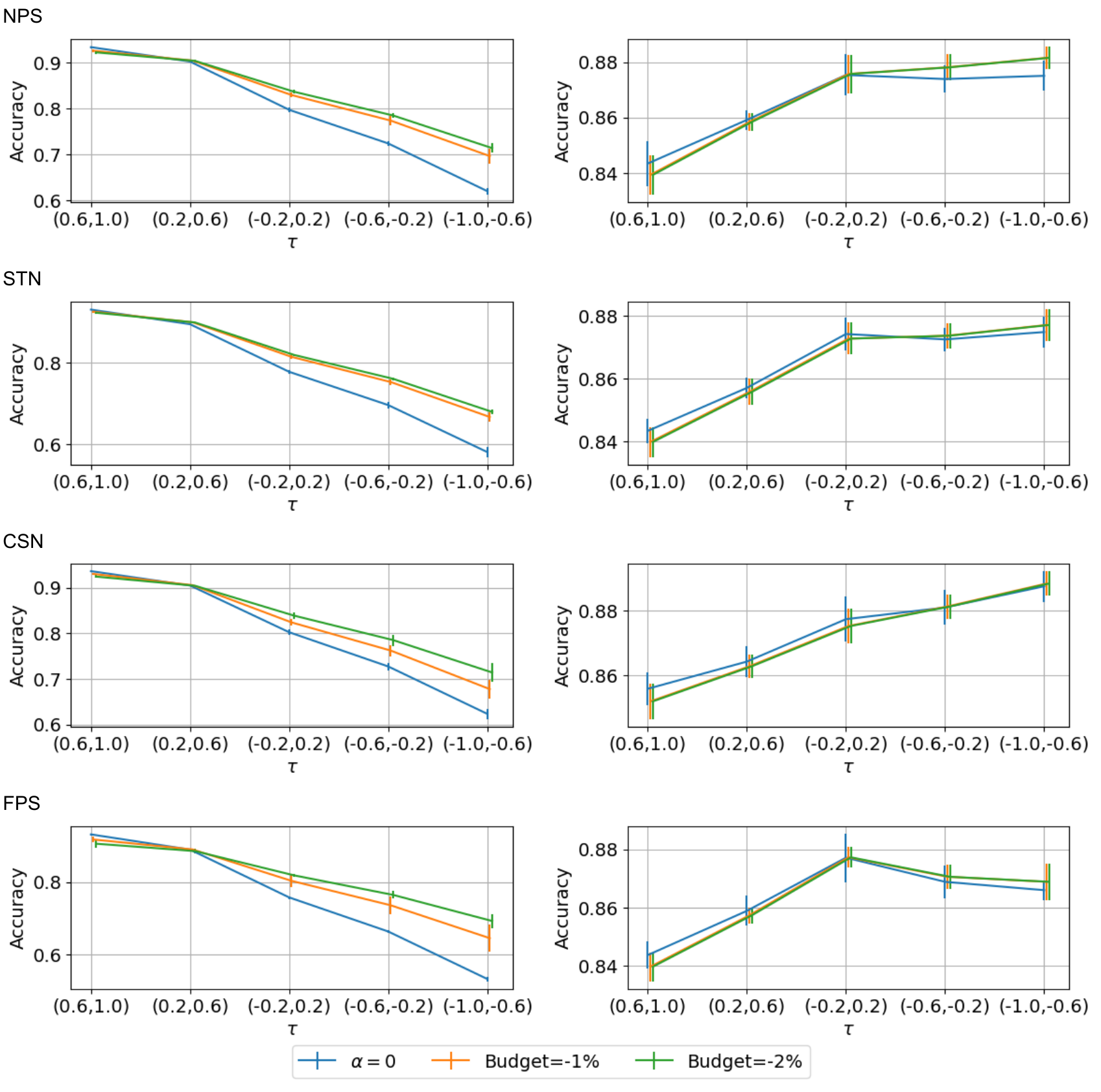}
    \caption{Attributes of people with long sleeves as the main task, and hat as the auxiliary task (left). The main and auxiliary tasks are exchanged in the right subfigures.}
    \label{fig:select_alpha,attributes_of_people,t=2,4}
\end{figure}

\begin{figure}[H]
    \centering
    \includegraphics[width=\textwidth]{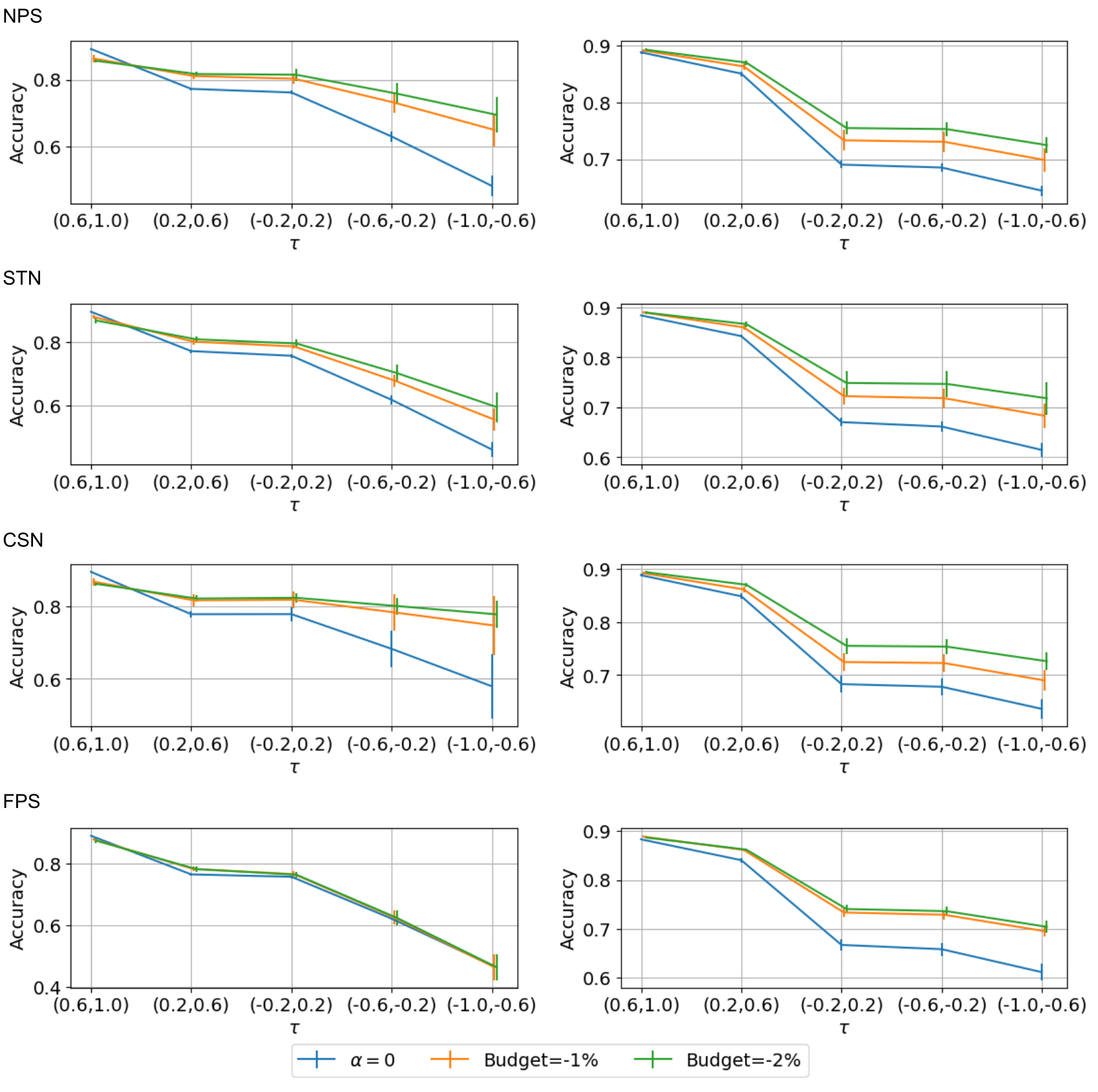}
    \caption{Attributes of people with long hair as the main task, and hat as the auxiliary task (left). The main and auxiliary tasks are exchanged in the right subfigures.}
    \label{fig:select_alpha,attributes_of_people,t=0,2}
\end{figure}

\begin{figure}[H]
    \centering
    \includegraphics[width=\textwidth]{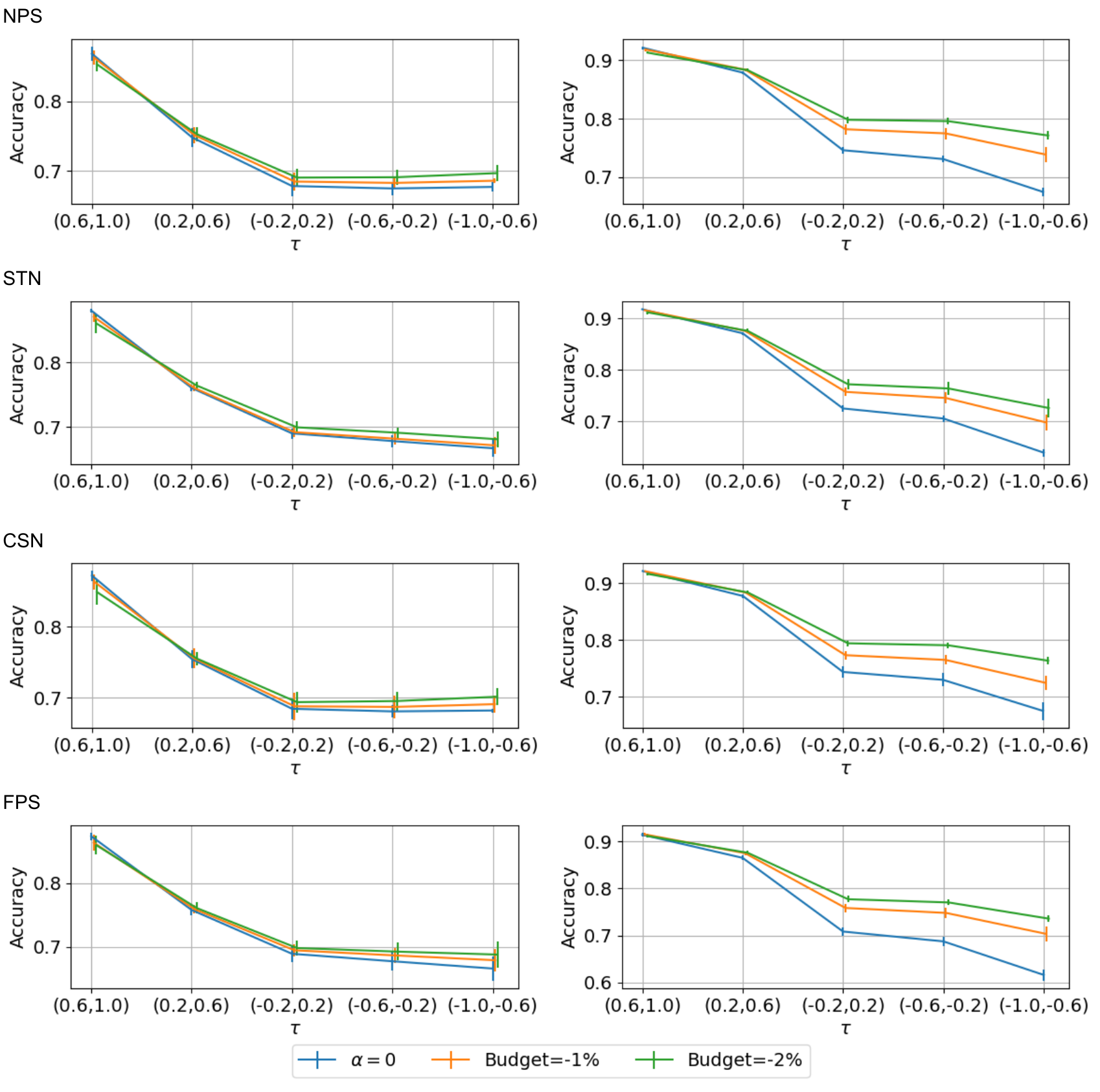}
    \caption{Attributes of people with glasses as the main task, and hat as the auxiliary task (left). The main and auxiliary tasks are exchanged in the right subfigures.}
    \label{fig:select_alpha,attributes_of_people,t=1,2}
\end{figure}

\begin{figure}[H]
    \centering
    \includegraphics[width=\textwidth]{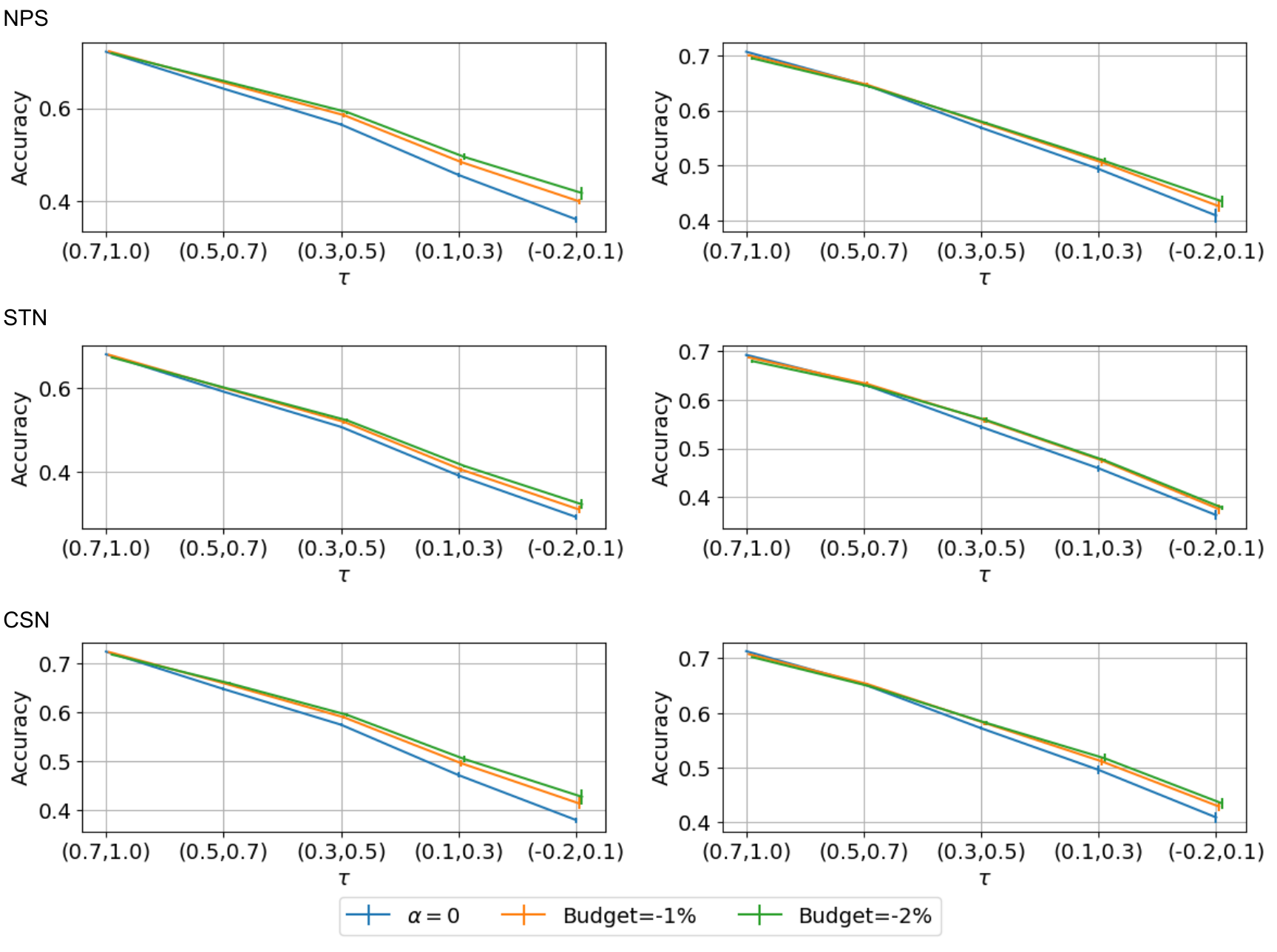}
    \caption{Taskonomy with object classification as the main task, and scene classification as the auxiliary task (left). The main and auxiliary tasks are exchanged in the right subfigures.}
    \label{fig:select_alpha,taskonomy}
\end{figure}

\subsection{Results with access to the optimal $\alpha$}
\label{sec:optimal alpha}

These are the same type of results shown in Section 6 of the main text, but for the remaining tasks across both datasets. For all tasks except the one mentioned in Section~\ref{sec:anomalous results}, the optimal alpha increases w.r.t. the severity of target shift, and there is a significant improvement in accuracy using the optimal alpha. These are the same observations that we made in the main text.

\begin{figure}[H]
    \centering
    \includegraphics[width=\textwidth]{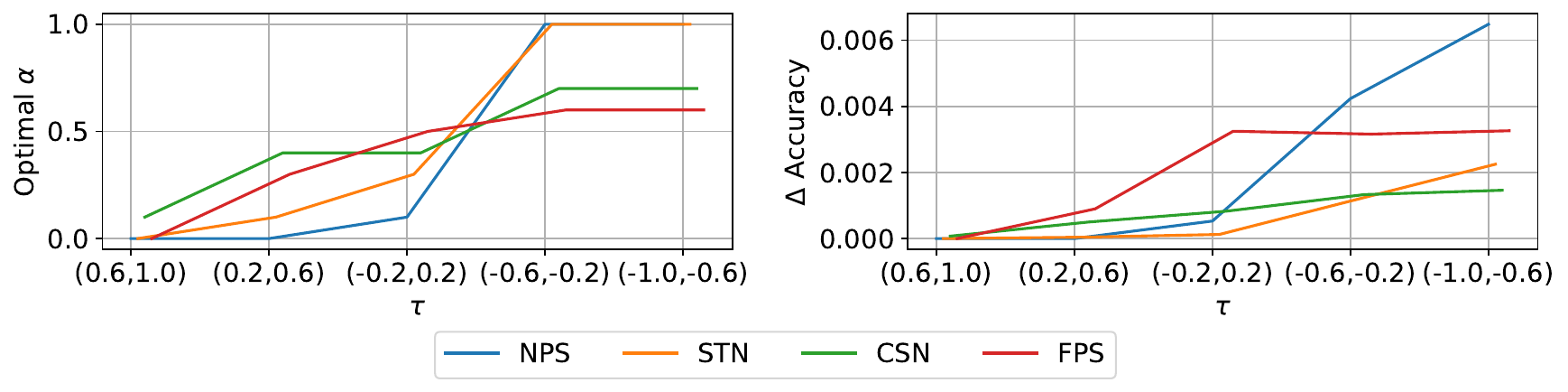}
    \caption{Attributes of people with long sleeves as the main task, and hat as the auxiliary task.}
    \label{fig:optimal_alpha,attributes_of_people,t=2,4,mt=1}
\end{figure}

\begin{figure}[H]
    \centering
    \includegraphics[width=\textwidth]{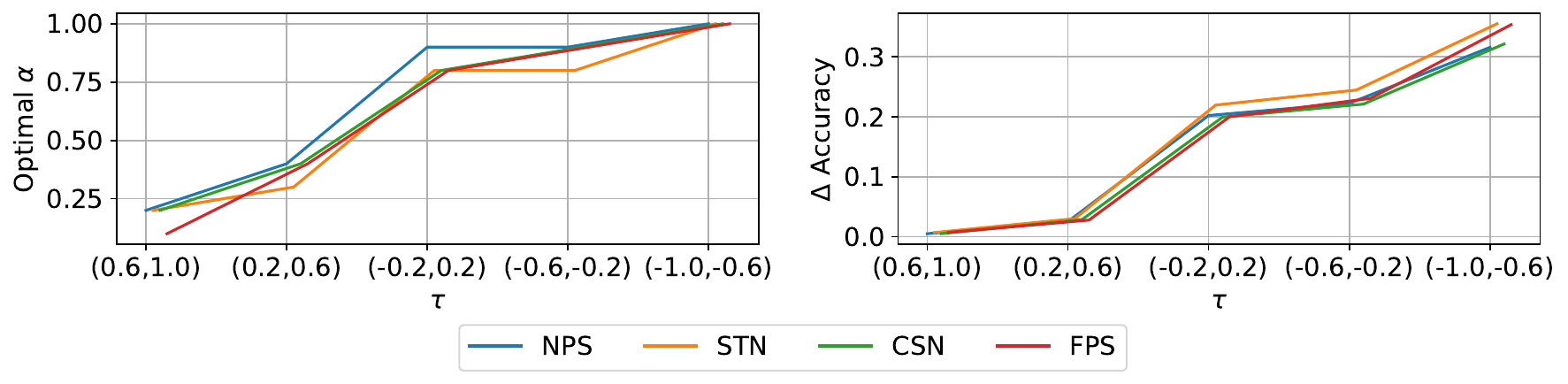}
    \caption{Attributes of people with long hair as the main task, and hat as the auxiliary task.}
    \label{fig:optimal_alpha,attributes_of_people,t=0,2,mt=0}
\end{figure}

\begin{figure}[H]
    \centering
    \includegraphics[width=\textwidth]{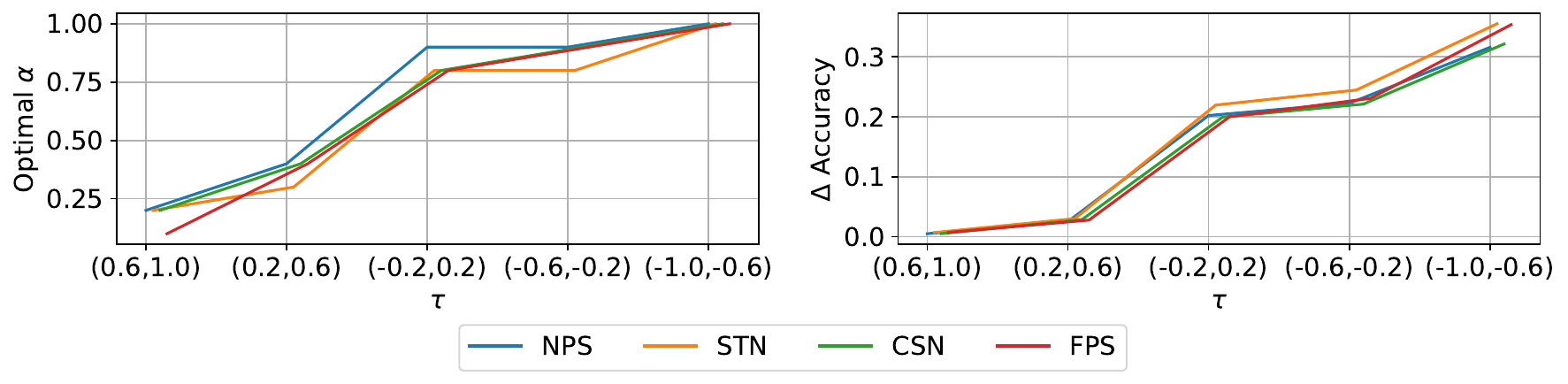}
    \caption{Attributes of people with hat as the main task, and long hair as the auxiliary task.}
    \label{fig:optimal_alpha,attributes_of_people,t=0,2,mt=1}
\end{figure}

\begin{figure}[H]
    \centering
    \includegraphics[width=\textwidth]{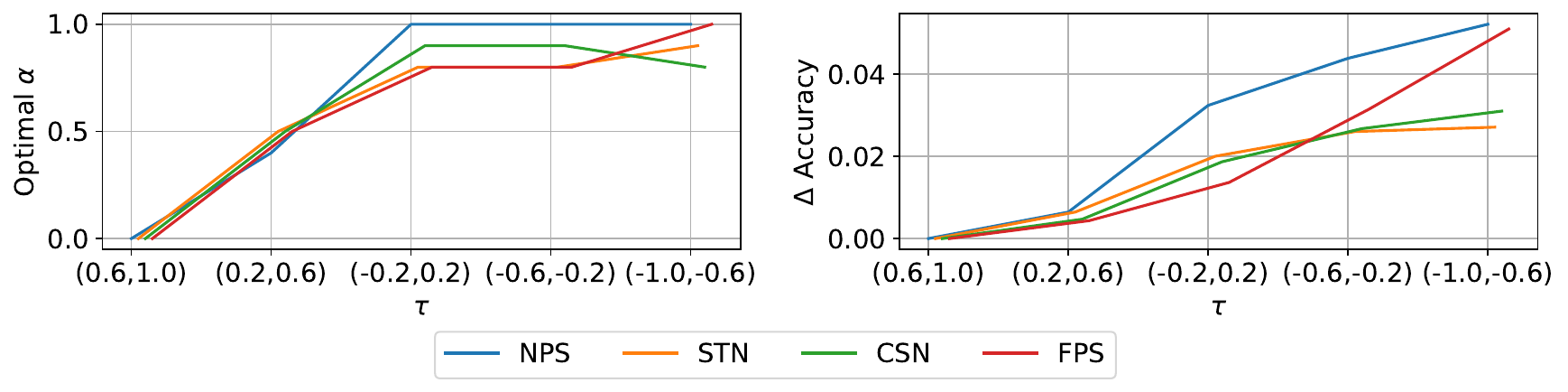}
    \caption{Attributes of people with glasses as the main task, and hat as the auxiliary task.}
    \label{fig:optimal_alpha,attributes_of_people,t=1,2,mt=0}
\end{figure}

\begin{figure}[H]
    \centering
    \includegraphics[width=\textwidth]{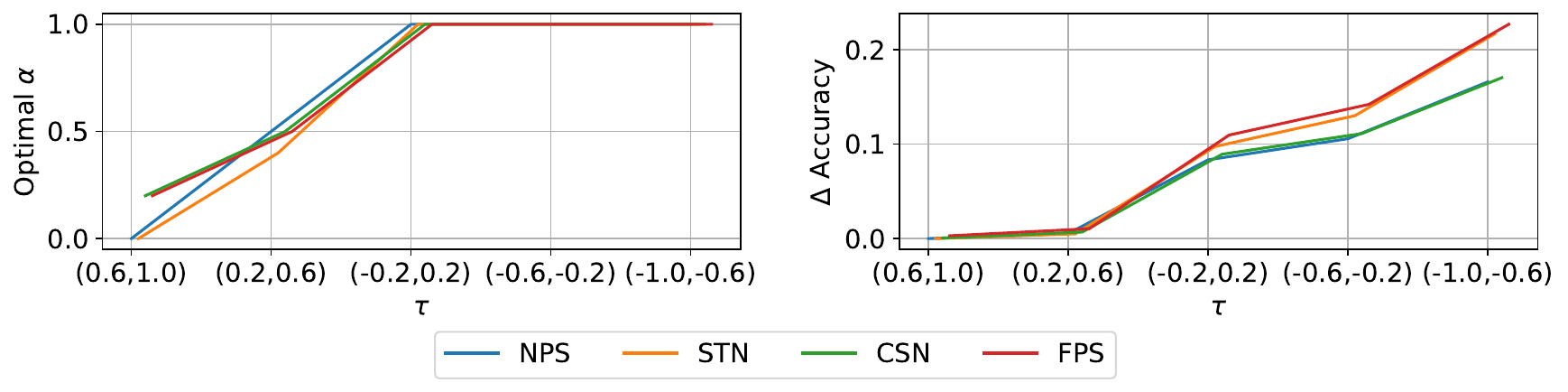}
    \caption{Attributes of people with hat as the main task, and glasses as the auxiliary task.}
    \label{fig:optimal_alpha,attributes_of_people,t=1,2,mt=1}
\end{figure}

\begin{figure}[H]
    \centering
    \includegraphics[width=\textwidth]{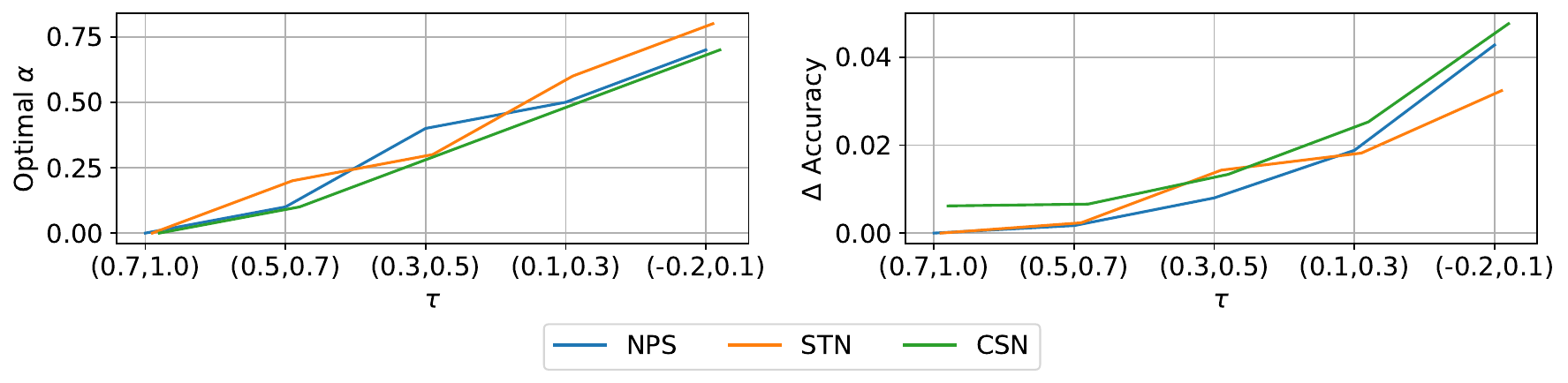}
    \caption{Taskonomy with scene classification as the main task, and object classification as the auxiliary task.}
    \label{fig:optimal_alpha,taskonomy,mt=1}
\end{figure}

\subsection{Results for a range of $\alpha$}
\label{sec:alpha range}

These results show the test accuracy for a range of $\alpha$, not just the optimal one. They can therefore be seen as a more granular view of the information presented in the left subfigures of Section~\ref{sec:optimal alpha}. In each set of four figures, each subfigure represents a different severity of target shift. Recall that the target shift is mildest when $\tau = 1$, and most severe when $\tau = -1$. These results support our interpretation of $\alpha$ as being a trade-off. That is, increasing $\alpha$ removes spurious dependencies that are predictive in the ID setting, and not in the out-of-distribution setting. Therefore, the optimal $\alpha$ tends to be small when the target shift is mild (top left), and large when the target shift is severe (bottom right). The optimal $\alpha$ tends to be somewhere in the middle when the target shift is in between being mild and severe. These conclusions hold consistently across all tasks, except for the one mentioned in Section~\ref{sec:anomalous results}.

\begin{figure}[H]
    \centering
    \includegraphics[width=0.85\textwidth]{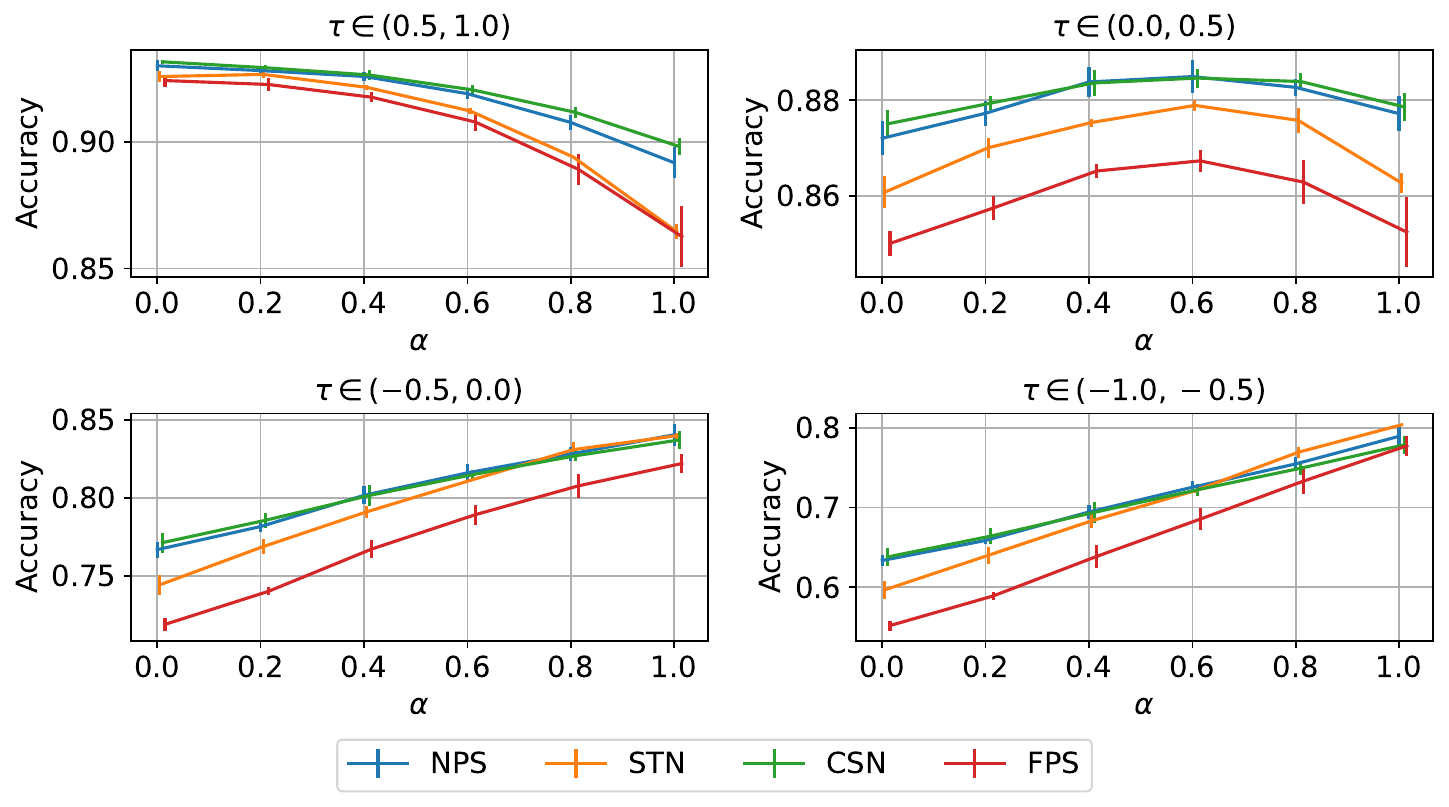}
    \caption{Attributes of people with hat as the main task, and long sleeves as the auxiliary task.}
    \label{fig:alpha_range,attributes_of_people,t=2,4,mt=0}
\end{figure}

\begin{figure}[H]
    \centering
    \includegraphics[width=0.85\textwidth]{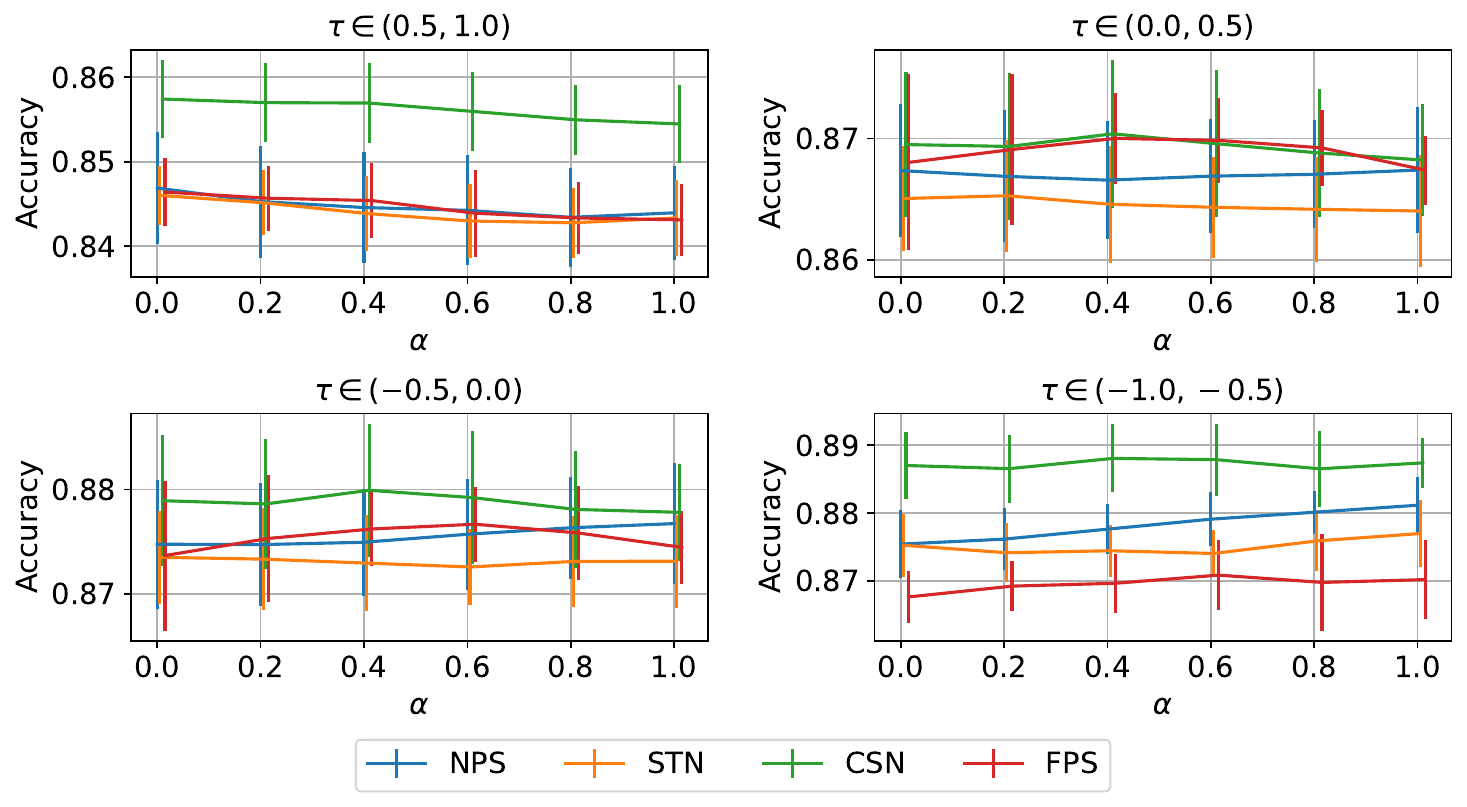}
    \caption{Attributes of people with long sleeves as the main task, and hat as the auxiliary task.}
    \label{fig:alpha_range,attributes_of_people,t=2,4,mt=1}
\end{figure}

\begin{figure}[H]
    \centering
    \includegraphics[width=0.85\textwidth]{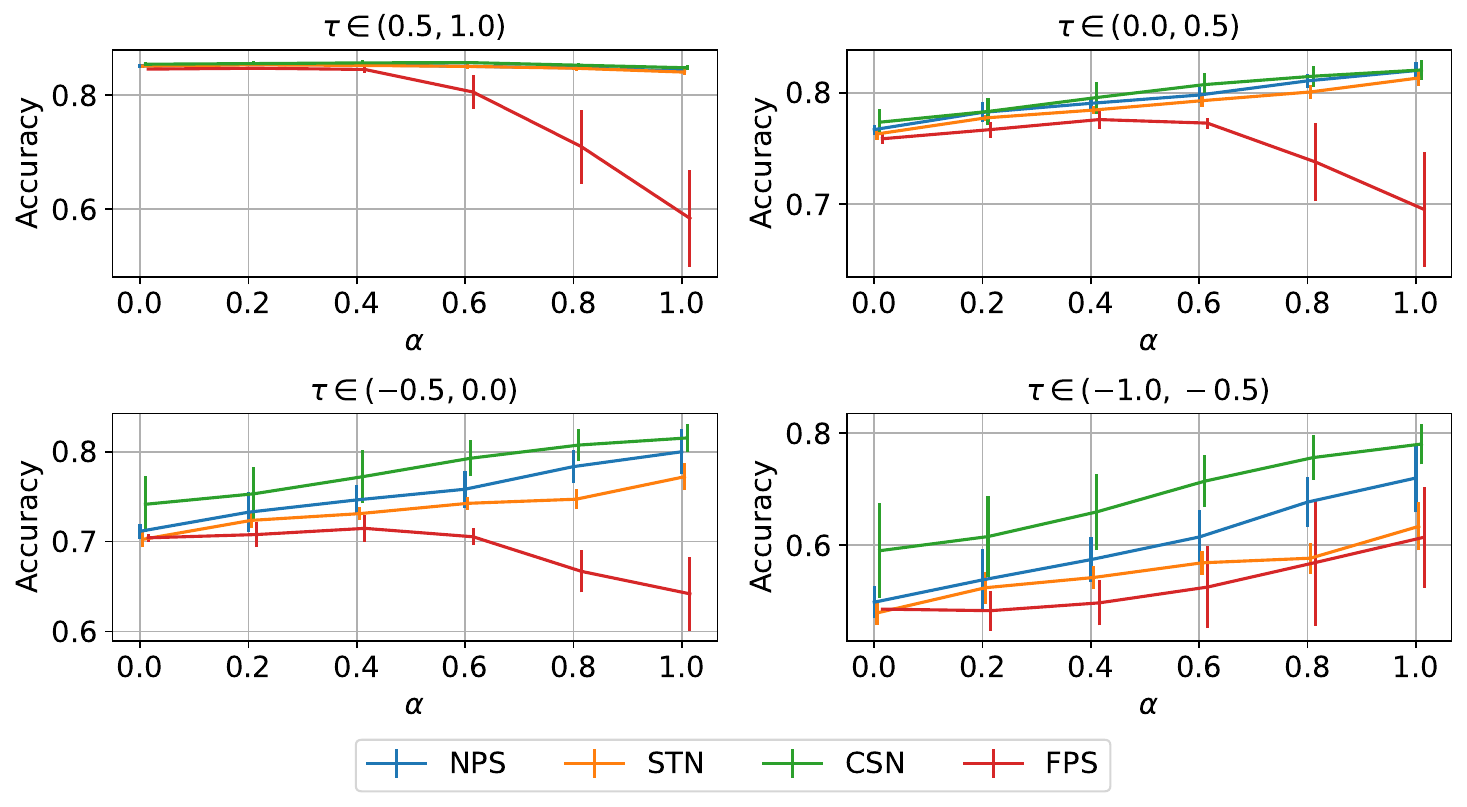}
    \caption{Attributes of people with long hair as the main task, and hat as the auxiliary task.}
    \label{fig:alpha_range,attributes_of_people,t=0,2,mt=0}
\end{figure}

\begin{figure}[H]
    \centering
    \includegraphics[width=0.85\textwidth]{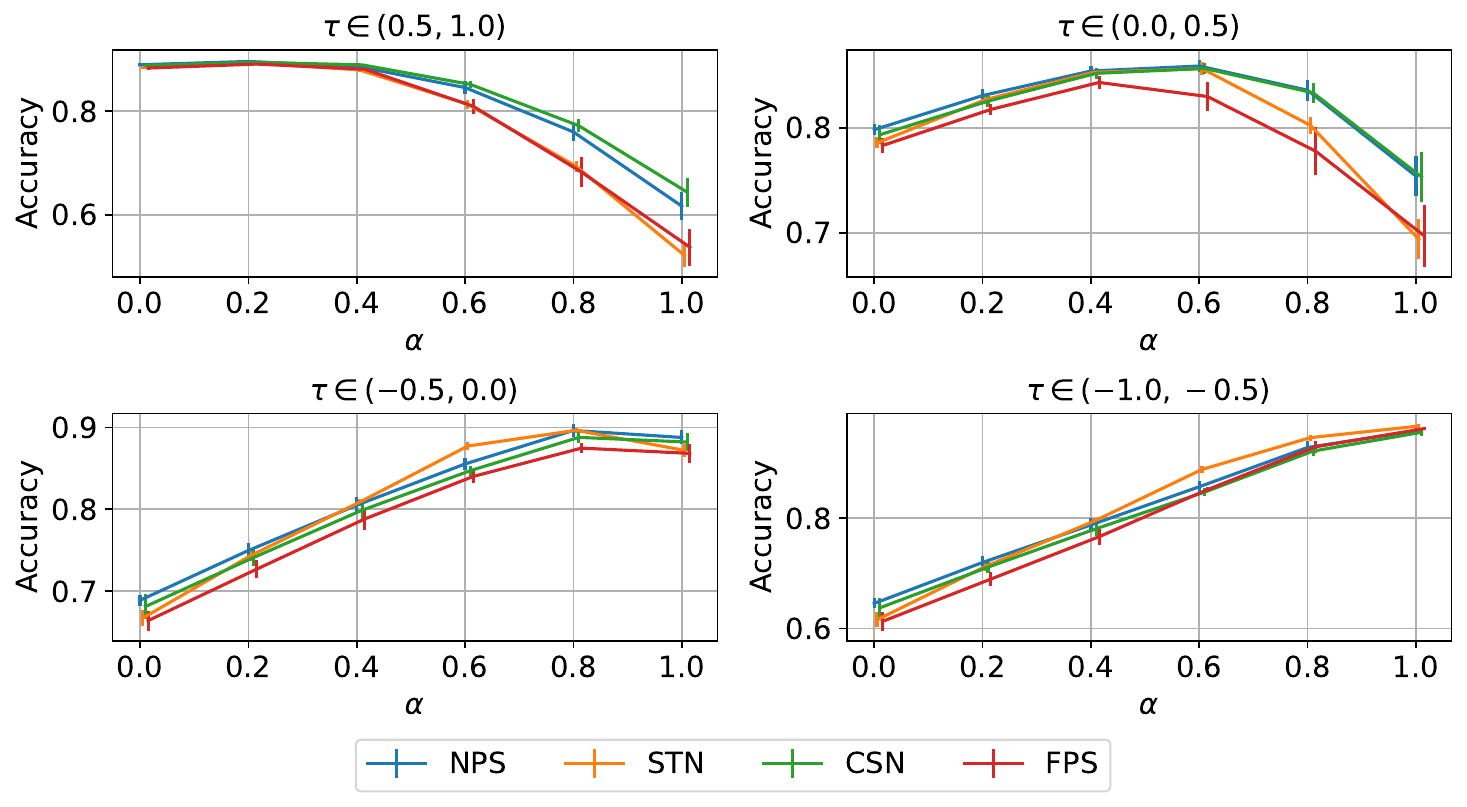}
    \caption{Attributes of people with hat as the main task, and long hair as the auxiliary task.}
    \label{fig:alpha_range,attributes_of_people,t=0,2,mt=1}
\end{figure}

\begin{figure}[H]
    \centering
    \includegraphics[width=0.85\textwidth]{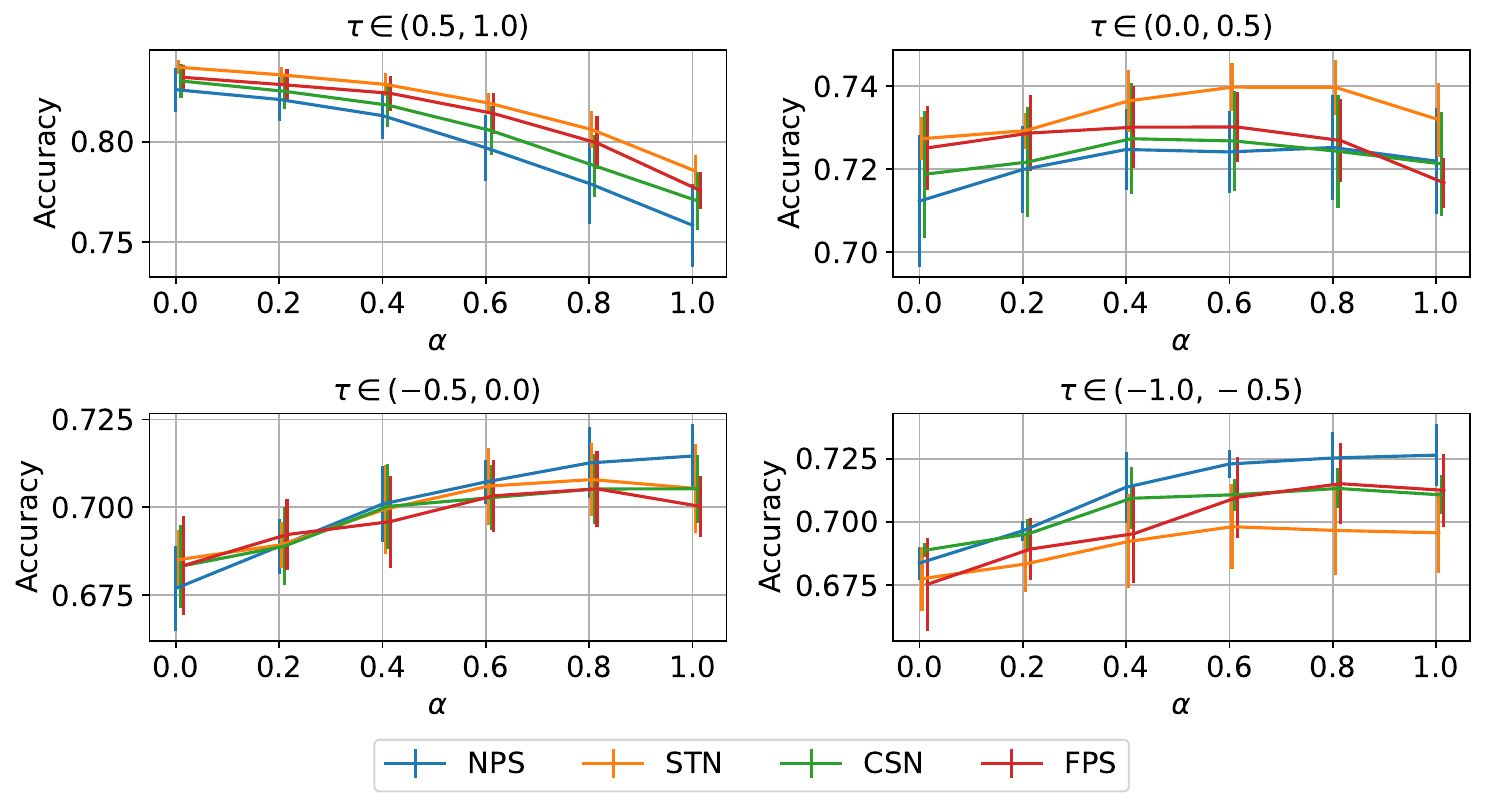}
    \caption{Attributes of people with glasses as the main task, and hat as the auxiliary task.}
    \label{fig:alpha_range,attributes_of_people,t=1,2,mt=0}
\end{figure}

\begin{figure}[H]
    \centering
    \includegraphics[width=0.85\textwidth]{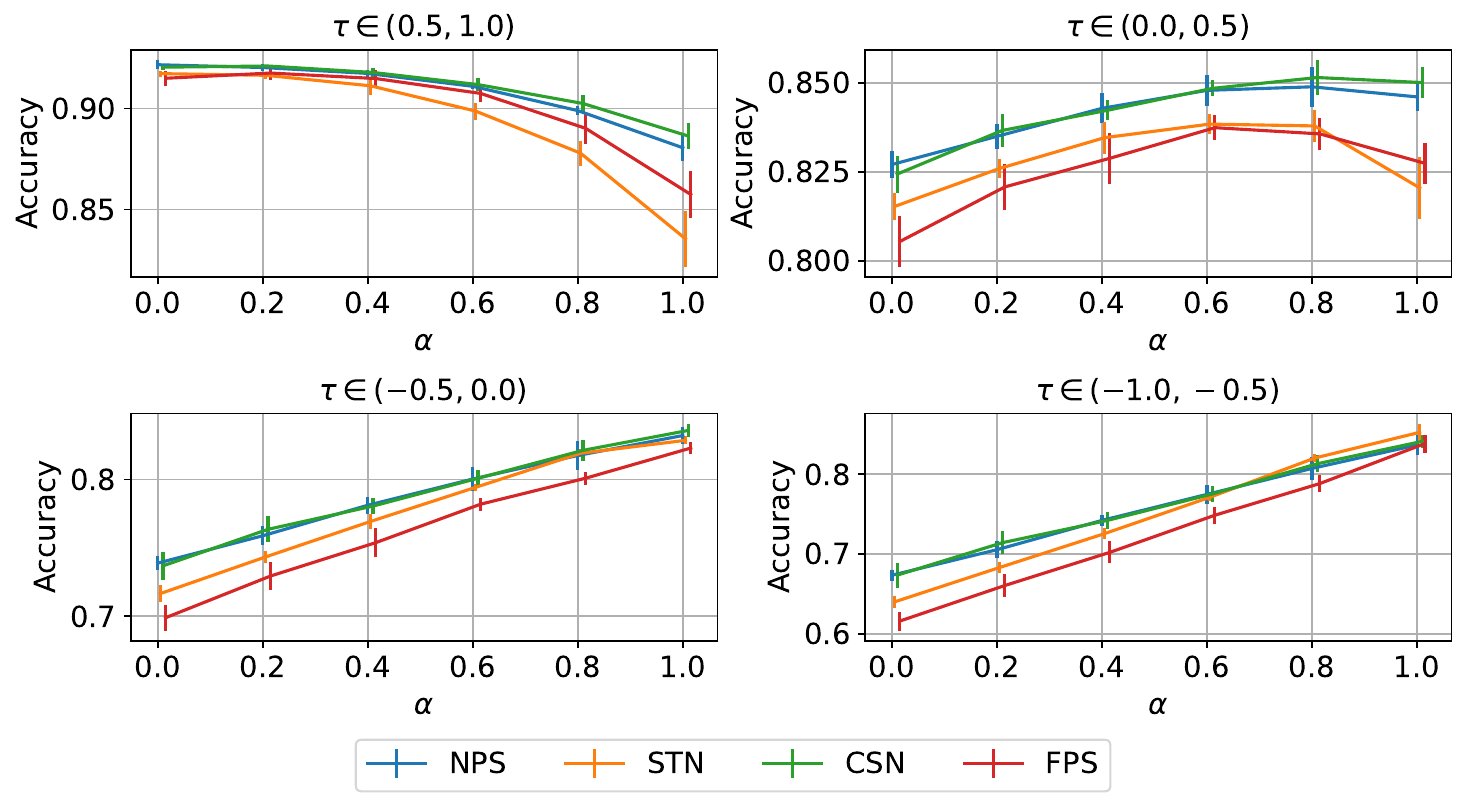}
    \caption{Attributes of people with hat as the main task, and glasses as the auxiliary task.}
    \label{fig:alpha_range,attributes_of_people,t=1,2,mt=1}
\end{figure}

\begin{figure}[H]
    \centering
    \includegraphics[width=0.85\textwidth]{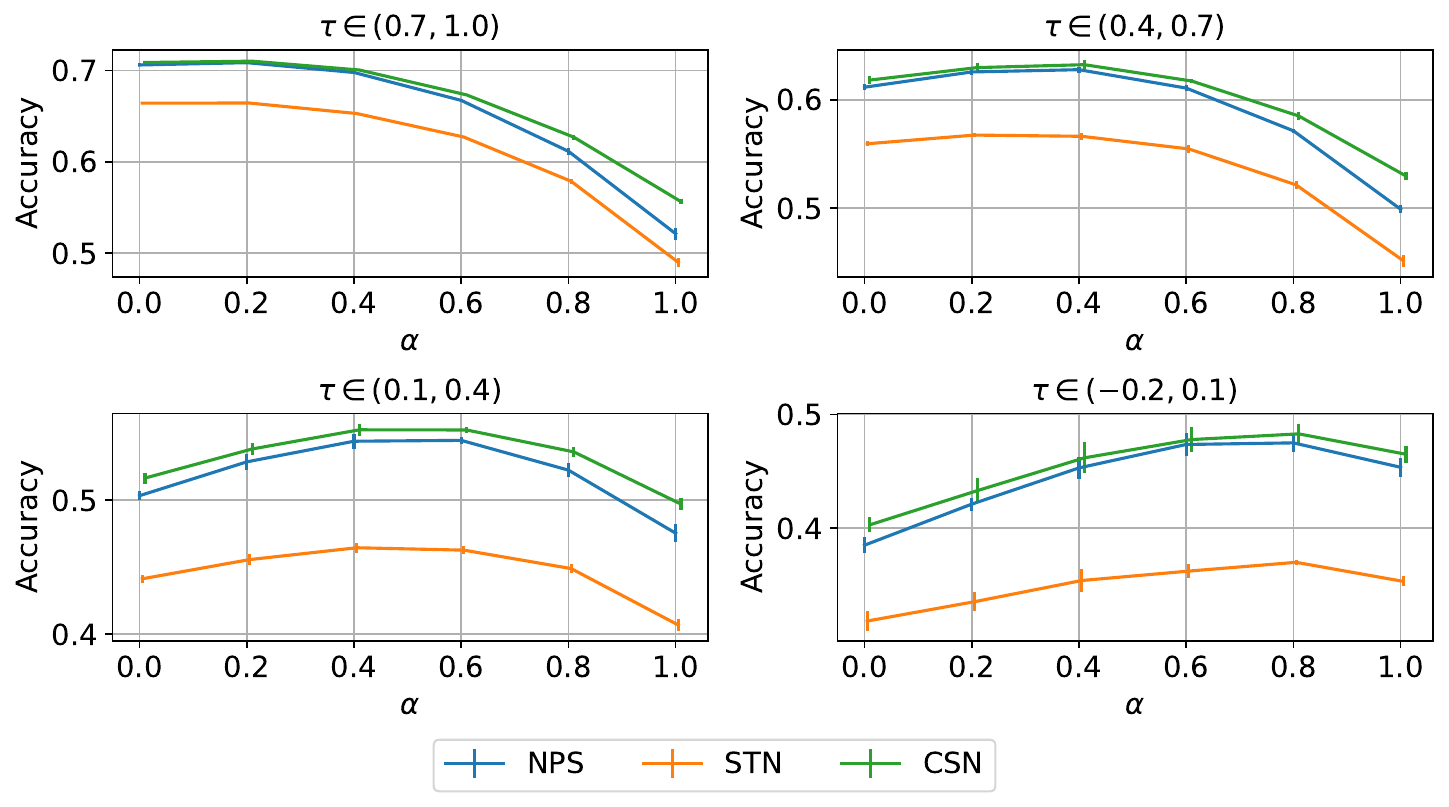}
    \caption{Taskonomy with object classification as the main task, and scene classification as the auxiliary task.}
    \label{fig:alpha_range,taskonomy,mt=0}
\end{figure}

\begin{figure}[H]
    \centering
    \includegraphics[width=0.85\textwidth]{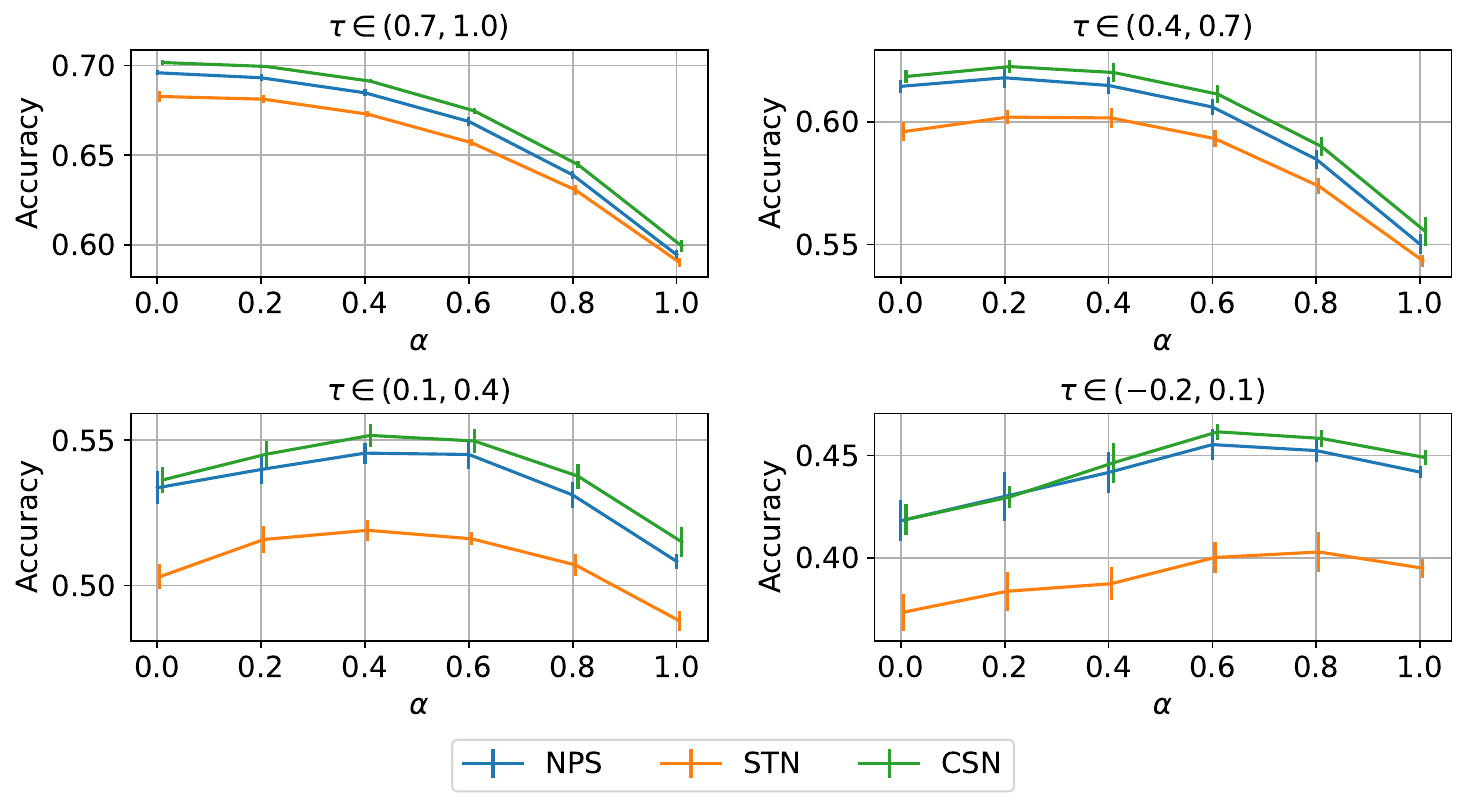}
    \caption{Taskonomy with scene classification as the main task, and object classification as the auxiliary task.}
    \label{fig:alpha_range,taskonomy,mt=1}
\end{figure}

\end{document}